\documentclass[letterpaper]{article} 
\usepackage{aaai25}  
\usepackage{times}  
\usepackage{helvet}  
\usepackage{courier}  
\usepackage[hyphens]{url}  
\usepackage{graphicx} 
\def\UrlFont{\rm}  
\usepackage{natbib}  
\usepackage{caption} 
\usepackage{algorithm}
\usepackage{algorithmic}

\usepackage{newfloat}
\usepackage{listings}
\DeclareCaptionStyle{ruled}{labelfont=normalfont,labelsep=colon,strut=off} 
\floatstyle{ruled}
\newfloat{listing}{tb}{lst}{}
\floatname{listing}{Listing}
\title{Amplifier: Bringing Attention to Neglected Low-Energy Components\\ in Time Series Forecasting}

\author{
    Jingru Fei\textsuperscript{\rm 1}, 
    Kun Yi\textsuperscript{\rm 2},
    Wei Fan\textsuperscript{\rm 3},
    Qi Zhang\textsuperscript{\rm 4},
    Zhendong Niu\textsuperscript{\rm 1}\thanks{Corresponding author}
}

\affiliations{
    \textsuperscript{\rm 1}Beijing Institute of Technology\\
    \textsuperscript{\rm 2}State Information Center of China\\
    \textsuperscript{\rm 3}University of Oxford\\
    \textsuperscript{\rm 4}Tongji University\\
    \{jingrufei, zniu\}@bit.edu.cn, kunyi.cn@gmail.com, weifan.oxford@gmail.com, zhangqi\_cs@tongji.edu.cn
}

\usepackage{bibentry}

\newtheorem{myDef}{Definition}
\newtheorem{myTheo}{Theorem}
\newtheorem{myLemma}{Lemma}
\usepackage{amsmath}
\usepackage{amssymb}
\usepackage{booktabs}
\usepackage{multirow}
\usepackage{subfigure}
\usepackage{xcolor}

\begin{document}

\maketitle


\begin{abstract}
We propose an \textit{energy amplification technique} to address the issue that existing models easily overlook low-energy components in time series forecasting. This technique comprises an \textit{energy amplification block} and an \textit{energy restoration block}. The energy amplification block enhances the energy of low-energy components to improve the model's learning efficiency for these components, while the energy restoration block returns the energy to its original level. Moreover, considering that the energy-amplified data typically displays two distinct energy peaks in the frequency spectrum, we integrate the energy amplification technique with a seasonal-trend forecaster to model the temporal relationships of these two peaks independently, serving as the backbone for our proposed model, \textit{Amplifier}. Additionally, we propose a  \textit{semi-channel interaction temporal relationship enhancement block} for Amplifier, which enhances the model's ability to capture temporal relationships from the perspective of the \textit{commonality} and \textit{specificity} of each channel in the data. Extensive experiments on eight time series forecasting benchmarks consistently demonstrate our model's superiority in both effectiveness and efficiency compared to state-of-the-art methods.

\end{abstract}

\begin{links}
    \link{Code}{https://github.com/aikunyi/Amplifier}
\end{links}

\section{Introduction}
Time series forecasting holds significant importance in real-life applications, encompassing various fields such as financial markets~\cite{yi2024deep}, weather forecasting~\cite{yifilternet}, traffic flow prediction~\cite{yu2018spatio,fan2022depts}, energy planning~\cite{yi2024fouriergnn}.
Recently, the rapid advancement of deep learning has given rise to various models for time series forecasting, including RNN-based methods (e.g., LSTNet~\cite{lai2018modeling}, DeepAR~\cite{salinas2020deepar}),  
TCN-based methods (e.g., SCINet~\cite{liu2022scinet}, TimesNet~\cite{wu2023timesnet}),
Transformer-based methods (e.g., PatchTST~\cite{nietime}, iTransformer~\cite{liuitransformer}), 
and Linear-based methods (e.g., DLinear~\cite{zeng2023transformers}, RLinear~\cite{li2023revisiting}), etc. 

Although these deep learning methods have demonstrated competitive performance across various scenarios, they possess several inherent drawbacks in their ability to learn from different energy components, and tend to \textbf{focus on learning high-energy components while neglecting low-energy components}, leading to an incomplete utilization of the informative aspects crucial for time series forecasting.

\textbf{What are low-energy components?} In signal processing and analysis, the energy of a signal refers to the total amount of signal strength or power~\cite{lathi1998signal}. Low-energy components in a signal refer to frequency components with smaller amplitudes within the frequency spectrum~\cite{oppenheim1999discrete}.

\begin{figure}[!t]
    \centering
    \includegraphics[width=1\linewidth]{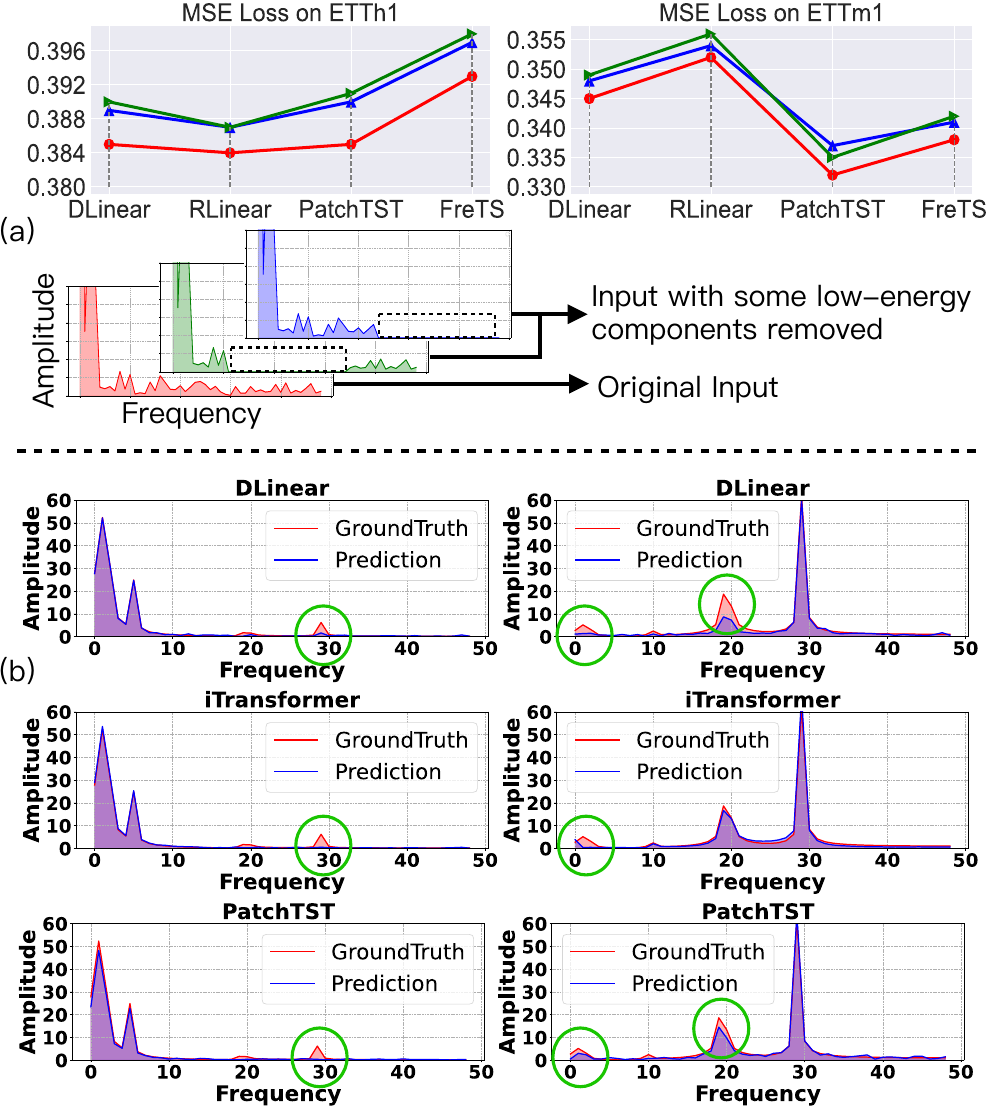}
    \vspace{-6mm}
    \caption{Analysis about low-energy components in time series forecasting. (a) \textit{Indispensability}: Discarding low-energy components results in an increased MSE value. 
    (b) \textit{Dependence on energy magnitude}: The components that are ignored are consistently those with low energy, irrespective of their position within the frequency bands.
    }
    \label{fig:motivation}
    \vspace{-6mm}
\end{figure}

\textbf{Low-energy components are indispensable.}
In contexts like signal compression or model simplification, low-energy components might be dismissed as unimportant or treated as noise. However, in scenarios where subtle changes or background patterns are significant, low-energy components can hold critical details. For example, in weather forecasting, small meteorological shifts can accumulate over time, leading to dramatic changes in weather patterns. In financial markets, minor fluctuations can trigger larger trends or volatility, with high-frequency trading systems often capitalizing on these small changes to execute profitable trades. We also have empirically validated the indispensability of low-energy components in time series forecasting. As illustrated in Figure \ref{fig:motivation}(a), directly filtering out low-energy components leads to increased MSE values across different types of models (PatchTST, RLinear, DLinear, FreTS), highlighting the crucial role of low-energy components in enhancing prediction accuracy.

\textbf{Low-energy components are often overlooked.}
We generate two signals with different energy distributions (corresponding to the left and right of Figure \ref{fig:motivation}(b)), and conduct iTransformer~\cite{liuitransformer}, PatchTST~\cite{nietime}, and DLinear~\cite{zeng2023transformers} on them respectively. As detailed in Figure \ref{fig:motivation}(b), the phenomenon of ignorance (rf. small green circles) always occurs with low-energy components, regardless of the frequency band in which they are located.

In this paper, to address the issue of low-energy components being neglected, we propose an \textbf{Energy Amplification Technique}, which comprises an \textit{energy amplification block} and an \textit{energy restoration block}. The energy amplification block is designed to boost the energy of low-energy components, enhancing the model's learning efficiency for these components, while the energy restoration block brings the energy back to its original level. 
This technique can be applied as a general approach in other time series forecasting models to enhance their performance.

To better harness the potential of the energy amplification technique, we design a new model, \textbf{Amplifier}, based on the characteristics of the data after energy amplification.
Specifically, since the data processed by the energy amplification block exhibits two energy peaks in the spectrum,
it is crucial to separate these two peaks to prevent them from simultaneously dispersing the attention of a single module or layer. Thankfully, the trend and seasonal components obtained from the seasonal-trend decomposition (abbreviated as STD) correspond directly to these two energy peaks. And the widespread use of STD reflects its strong acceptance among researchers. Thus we integrate the energy amplification technique with a seasonal-trend forecaster (based on STD) to model the temporal relationships for these two peaks separately, forming the backbone of Amplifier. To further improve information utilization, we also develop a \textit{semi-channel interaction temporal relationship enhancement block} (abbreviated as SCI block) as an optional built-in component for Amplifier to enhance the performance from the perspective of the \textit{commonality} and \textit{specificity} of each channel in time series data. 
 
Our contributions can be summarized as follows:
\begin{itemize}
    \item We identify a common issue in existing time series forecasting models, i.e., the neglect of low-energy components, which can lead to performance degradation.
    \item We propose an energy amplification technique to address the above issue, serving as a general method to enhance the performance of other foundational models.
    \item To better leverage the energy amplification technique, we combine this technique with a temporal relationship enhancement and a seasonal-trend forecaster to propose a novel model for time series forecasting, called Amplifier. 
    \item Extensive experiments on 8 real-world datasets demonstrate that our Amplifier consistently outperforms state-of-the-art methods while offering higher efficiency.
\end{itemize}

\section{Related Work}
\subsection{Deep Time Series Forecasting}
With the rapid advancement of deep learning technology and the increasing importance of time series forecasting, various models have emerged in a flourishing and competitive landscape~\cite{Lim_2021}. 
Recent studies have proposed a series of upgraded transformer-based models for time series forecasting, such as LogTrans~\cite{li2019enhancing}, Informer~\cite{zhou2021informer}, and Pyraformer~\cite{liu2021pyraformer}. Meanwhile, some models focus on functional improvements to the Transformer architecture, making it more suitable for time series forecasting tasks, including Autoformer~\cite{wu2021autoformer}, FEDformer~\cite{zhou2022fedformer}, PatchTST~\cite{nietime}, and iTransformer~\cite{liuitransformer}.
In addition to the Transformer architecture, lightweight and efficient MLP architectures have also been favored by many researchers. The brilliant debut of DLinear~\cite{zeng2023transformers} paved the way for MLP models to shine in the field of time series forecasting, leading to the emergence of numerous MLP models, such as: RLinear~\cite{li2023revisiting}, TiDE~\cite{das2023long}, and SparseTSF~\cite{lin2024sparsetsf}. 

\subsection{Advancements in Frequency Domain Techniques}
Recent studies have increasingly leveraged frequency techniques to enhance both the accuracy and efficiency of time series forecasting~\cite{kunyi_2023_survey}. FEDformer~\cite{zhou2022fedformer} achieves a fast attention mechanism through low-rank approximated transformation in the frequency domain. FITS~\cite{xu2023fits} uses frequency domain interpolation to make predictions and essentially functions as a low-pass filter. FreTS~\cite{yi2024frequency} retains the simplicity and efficiency of MLP architecture while incorporating the global perspective and energy aggregation characteristics of the frequency domain. 
Although the above methods demonstrate that frequency domain techniques hold great potential for time series forecasting tasks, they typically either treat components with different energy levels uniformly or focus exclusively on low-frequency components (high-energy parts) while discarding high-frequency components (low-energy parts). This oversight often leads to sub-optimal performance in time series forecasting.

More recently, a few works have begun to address these limitations. Fredformer~\cite{piao2024fredformer} mitigates frequency bias in the Transformer architecture by proposing a framework that learns features equally across different frequency bands. However, Fredformer primarily explores the Transformer architecture and the improvements are specific to that model.  In contrast, in this paper, we propose a unified method applicable to the main paradigm networks, including Transformer and MLP, which employs energy amplification technique to bring model's attention to neglected low-energy components in time series forecasting, fully leveraging all available data to achieve superior performance. Moreover, compared to Fredformer, Amplifier offers much faster training speeds and a significantly smaller parameter scale.

\section{Preliminaries}
In this section, we conduct a thorough analysis of the phenomenon where low-energy components are ignored in time series forecasting as mentioned in Introduction section.

\paragraph{Energy} Energy can be referred to as a concept in signal processing that captures the overall strength or magnitude of a signal. Given a multivariate time series data $X\in \mathbb{R}^{C\times L}$ with the channel number of $C$ and the look-back window size of $L$, for the $i$-th time series data $X^i \in \mathbb{R}^{1\times L}$, its energy $\mathcal{E}(\mathcal{X}^i)$ can be defined as $ {\textstyle \sum_{n=0}^{L-1}} \left | \mathcal{X}[n] \right |^2 $ where $\mathcal{X}[n] \in \mathbb{C}^L$ is the Discrete Fourier Transform (DFT) of $X^i$. 
The spectrum of real-world datasets often shows a clear distinction between high and low energy levels, as illustrated by the ETTm1 dataset in the right half of Figure ~\ref{fig:motivation}(a).
The left side of the spectrum represents high-energy frequency points, while the right side represents low-energy frequency points.
We refer to a set of low-energy frequency points as low-energy components $\mathcal{X}^{i}_L$, and a set of high-energy frequency points as high-energy components  $\mathcal{X}^{i}_H$, where $\mathcal{E}(\mathcal{X}^{i}_H) \gg \mathcal{E}(\mathcal{X}^{i}_L)$.
Then, the $\mathcal{X}^{i}$ can be expressed as the sum of these components: 
$\mathcal{X}^i = \{\mathcal{X}^{i}_H, \mathcal{X}^{i}_L\}$.

\paragraph{Time Series Forecasting} For the multivariate time series input data $X\in \mathbb{R}^{C\times L}$ and its next $\tau$ data $Y\in \mathbb{R}^{C\times \tau}$, the time series forecasting task is to predict the next $\tau$ timestamps values $\hat{Y} \in \mathbb{R}^{C\times \tau}$ based on the historic data $X$ through a neural network $F_\theta$ parameterized by $\theta$. Then, the loss in time domain can be formulated as $\mathcal{L}(Y,\hat{Y};\theta)  = \left \| Y-\hat{Y}\right \|_2^2$. Since the data can be divided into high-energy components and low-energy components in frequency domain, the above loss function can be rewritten as an equivalent form in the frequency domain as below:
\begin{equation}
    \mathcal{L}(\mathcal{Y},\hat{\mathcal{Y}};\Theta) = \mathcal{L}(\mathcal{Y}_{H},\hat{\mathcal{Y}}_{H};\Theta_{H}) + \mathcal{L}(\mathcal{Y}_{L},\hat{\mathcal{Y}}_{L};\Theta_{L}),
\end{equation}
where $\mathcal{Y}$ and $\mathcal{\hat{Y}}$ is the DFT of $Y$ and $\hat{Y}$ respectively, and $\Theta$ denote parameters representing the characteristics in the frequency domain. 
To further investigate how the low-energy components are ignored in the learning process, we examine the loss function and parameter updates from the energy perspective, and we identify two main factors that lead to the neglect of low-energy components.

\begin{myTheo}\label{theorem_1}
In the initial stage of network training, the  loss of high-energy components $\mathcal{L}(\mathcal{Y}_{H},\mathcal{\hat{Y}}_{H};\Theta_{H})$ occupies a significantly larger proportion of the overall loss compared to the loss of low-energy components $\mathcal{L}(\mathcal{Y}_{L},\mathcal{\hat{Y}}_{L};\Theta_{L})$, that is:
\begin{equation}
    \frac{\mathcal{L}(\mathcal{Y}_{H},\hat{\mathcal{Y}}_{H};\Theta_{H})}{\mathcal{L}(\mathcal{Y},\mathcal{\hat{Y}};\Theta)} \gg \frac{\mathcal{L}(\mathcal{Y}_{L},\mathcal{\hat{Y}}_{L};\Theta_{L})}{\mathcal{L}(\mathcal{Y},\mathcal{\hat{Y}};\Theta)}.
\end{equation}
\end{myTheo}
We include the proof in Appendix \ref{proof_appendix}. It implies that in the initial stage of neural network training, the loss value of the high-energy components constitutes the majority of the total loss value. In other words, correcting the loss for the high-energy components can lead to a significant reduction in the total loss value and faster convergence from a global perspective. This guides the model to focus on reducing the error of the high-energy components during training, potentially leading to the neglect of the low-energy components.

\begin{myTheo}\label{theorem_2}
Parameter updates are influenced by the energy of their corresponding components, meaning that the updates for parameters $\Theta_L$ related to low-energy components are much less efficient than those $\Theta_H$ for high-energy components, which can be expressed as:
\begin{equation}
\frac{\mathcal{L}(\mathcal{Y}_{H},\mathcal{\hat{Y}}_{H};\Theta_{H})}{\partial \Theta_H} \gg \frac{\mathcal{L}(\mathcal{Y}_{L},\mathcal{\hat{Y}}_{L};\Theta_{L})}{\partial \Theta_L}.
\end{equation}
\end{myTheo}
The proof is shown in Appendix \ref{proof_appendix}. It indicates that during the training process, the parameters associated with low-energy components update more slowly and less efficiently than those associated with high-energy components. As a result, by the end of training, the parameters associated with low-energy components are further from their true values compared to those associated with high-energy components, indicating that these parameters have minimal participation during training, making them appear "ignored".

\section{Methodology}
As mentioned earlier, previous work typically ignores low-energy components, which are essential for accurate time series forecasting. To address this issue, we propose an \textit{energy amplification technique}, which can be applied as a general method to enhance the performance of existing forecasting models. Then, based on the technique, we propose a simple yet effective model, \textit{Amplifier}, for time series forecasting.

\begin{figure*}[!t]
    \centering
    \includegraphics[width=1\linewidth]{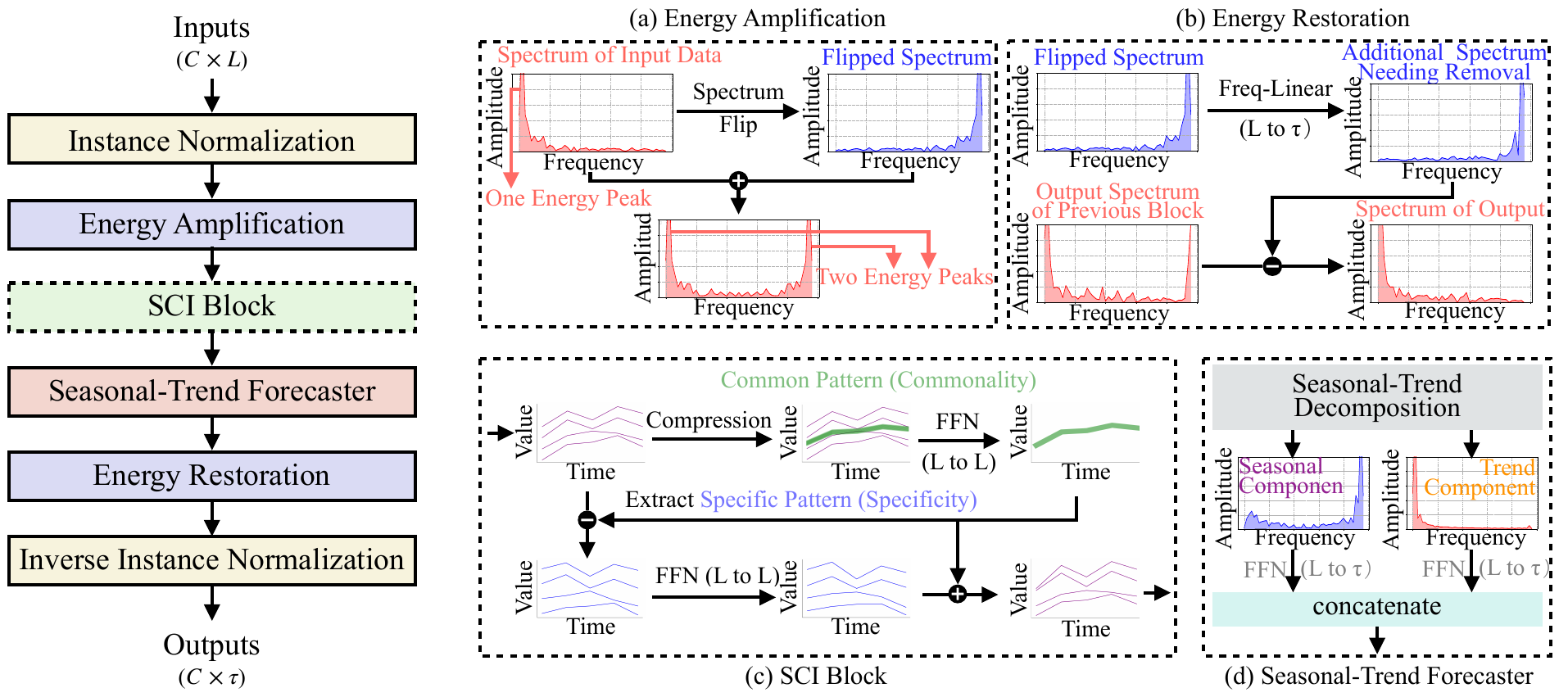}
    \vspace{-6mm}
    \caption{The overall architecture of Amplifier.  (i) The energy amplification block aims to increase the energy of low-energy components, while energy restoration block performs the inverse operation of the energy amplification block; (ii) SCI block is employed to capture both the common temporal pattern and specific temporal pattern; (iii) Seasonal-Trend Forecaster is then utilized to decompose seasonal and trend information, and then make predictions.
    }
    \label{fig:framework}
    \vspace{-4mm}
\end{figure*}

\subsection{Energy Amplification Technique}
The energy amplification technique consists of the \textit{energy amplification block} and the \textit{energy restoration block}. The energy amplification block aims to increase the energy of low-energy components to improve the model's learning efficiency for these components, while the energy restoration block restores the energy to its original level. 

\paragraph{Energy Amplification Block}
To enable the model to learn both low-energy components and high-energy components without bias, we aim to equalize the energy between them. We achieve this by transferring high energy from the low-frequency region to the low-energy components located in the high-frequency region through spectrum flipping. 

Energy can be represented by the square of its amplitude. For the multivariate time series input
data $X \in \mathbb{R}^{C\times L}$, its energy can be calculated by:
\begin{equation}
    \mathcal{E}(X) = \sum_{k=0}^{L-1} {\left | \mathcal{X}[k]  \right | }^{2},
\end{equation}
where $\mathcal{X} \in \mathbb{C}^{C\times L}$ is the DFT of $X$ and $k$ is the frequency point.
We shift high energy from the low-frequency region to the low-energy components located in the high-frequency region by flipping the spectrum, formulated as follows:
\begin{equation}\label{flip}
    \mathcal{X'} [k] = \mathcal{X} [T-k],
\end{equation}
where $\mathcal{X'} \in \mathbb{C}^{C\times L}$ refers to the flipped spectrum. By adding $\mathcal{X}$ and $\mathcal{X'}$ together and performing Inverse Discrete Fourier Transform (IDFT), we obtain the output $X_{\text{Amp}}$ of this block as detailed below:
\begin{equation}\label{eq:x_amp}
\begin{split}
    {\mathcal{X}}_{\text{Amp}}  &= \mathcal{X} + \mathcal{X'}, \\
    X_{\text{Amp}} &= \operatorname{IDFT}({\mathcal{X}}_{\text{Amp}}).
\end{split}
\end{equation}
At this point, the original low-energy components have gained energy comparable to the high-energy components, effectively bringing attention to neglected low-energy components, as shown in the following equation:
\begin{equation}
    {\mathcal{E}}_{\text{Amp}}[k] = {\mathcal{E}}_{\text{Amp}}[T-k].
\end{equation}

\paragraph{Energy Restoration Block}
The block is employed to remove the flipped spectrum added by the energy amplification block, serving as the inverse operation of energy amplification. First, we adjust the input length to align the prediction length by a frequency-domain linear operations as:
\begin{equation}\label{eq:freq_upsampling}
    \mathcal{Y'} = \mathcal{X'}\mathcal{W} +\mathcal{B}
\end{equation}
where $\mathcal{X'} \in \mathbb{C}^{C\times L}$ is the flipped spectrum, $\mathcal{W} \in \mathbb{C}^{L\times \tau}$ is a complex number weight matrix, $\mathcal{B} \in \mathbb{C}^{\tau}$ is a complex number bias, and $\mathcal{Y'} \in \mathbb{C}^{C\times \tau}$ denotes the additional spectrum needing removal. Then, we convert the output $Y_{\text{Amp}}$ of the previous block from the time domain to the frequency domain, remove the additional spectrum $\mathcal{Y'}$, and perform domain conversion to obtain the final prediction result:
\begin{equation}
\begin{split}
    {\mathcal{Y}}_{\text{Amp}} &= \operatorname{DFT}(Y_{\text{Amp}}), \\
    \mathcal{Y} &= {\mathcal{Y}}_{\text{Amp}} - \mathcal{Y'}, \\ 
    \hat{Y} &= \operatorname{IDFT}(\mathcal{Y}).
\end{split}
\end{equation}
For more explanation on the energy amplification technique, please refer to the Appendix \ref{explanation_appdenix}.

\subsection{Amplifier}
Given that the energy-amplified data usually exhibits two energy peaks, we combine the energy amplification technique with a seasonal-trend forecaster to model the temporal relationships for these two peaks separately, forming the backbone of our newly proposed model, {Amplifier}. 
The overall structure of Amplifier is illustrated in Figure \ref{fig:framework}, including four main blocks: energy amplification block, semi-channel interaction temporal relationship enhancement block (abbreviated as SCI block), seasonal-trend forecaster block, and energy restoration block. 
At the same time, we use Instance Normalization and its inverse operation to address non-stationarity in time series data
As the energy amplification technique has already been introduced in previous subsection, we describe the rest of the other blocks below.

\begin{table*}[h]
    \centering
    {\fontsize{10pt}{10pt}\selectfont
    \vspace{-1mm}
    \scalebox{0.68}{
    \begin{tabular}{l | c|c c|c c|c c| c c|c c |c c |c c |c c |c c| c c}
    \toprule[2pt]
       \multicolumn{2}{c|}{Models}  &\multicolumn{2}{c|}{\textbf{Amplifier}} &\multicolumn{2}{c|}{RLinear} &\multicolumn{2}{c|}{DLinear} &\multicolumn{2}{c|}{FreTS}  &\multicolumn{2}{c|}{FITS} &\multicolumn{2}{c|}{Fredformer} &\multicolumn{2}{c|}{iTransformer} &\multicolumn{2}{c|}{PatchTST}  &\multicolumn{2}{c|}{Stationary} &\multicolumn{2}{c}{TimesNet}  \\
       \cmidrule(r){1-2}\cmidrule(lr){3-4}\cmidrule(lr){5-6}\cmidrule(lr){7-8}\cmidrule(lr){9-10}\cmidrule(lr){11-12}\cmidrule(lr){13-14}\cmidrule(lr){15-16}\cmidrule(lr){17-18}\cmidrule(lr){19-20}\cmidrule(lr){21-22}
       \multicolumn{2}{c|}{Metrics}&MSE &MAE &MSE &MAE &MSE &MAE &MSE &MAE &MSE &MAE &MSE &MAE &MSE &MAE &MSE &MAE &MSE &MAE &MSE &MAE \\
       \midrule[1pt]
        \multicolumn{2}{c|}{ETTm1} &\textcolor{red}{\textbf{0.381}}  &\textcolor{red}{\textbf{0.394}}  &0.414 &0.407 &0.403 &0.407 &0.408  &0.416  &0.415  &0.408 &\textcolor{blue}{\textbf{0.384}}  &\textcolor{blue}{\textbf{0.395}} &0.407 &0.410 &0.387 &0.400 &0.481 &0.456 &0.400 &0.406 \\
        \cmidrule(lr){1-22}
        \multicolumn{2}{c|}{ETTm2} &\textcolor{red}{\textbf{0.276}}  &\textcolor{red}{\textbf{0.323}}  &0.286 &0.327 &0.350 &0.401 &0.321  &0.368  &0.286  &0.328 &\textcolor{blue}{\textbf{0.279}} &\textcolor{blue}{\textbf{0.324}} &0.288 &0.332 &0.281 &0.326 &0.306 &0.347 &0.291 &0.333 \\
        \cmidrule(lr){1-22}
        \multicolumn{2}{c|}{ETTh1} &\textcolor{red}{\textbf{0.430}}  &\textcolor{blue}{\textbf{0.428}}  &0.446 &0.434 &0.456 &0.452 &0.475  &0.463  &0.451  &0.440  &\textcolor{blue}{\textbf{0.435}} &\textcolor{red}{\textbf{0.426}}  &0.454 &0.447 &0.469 &0.454 &0.570 &0.537 &0.458 &0.450 \\
        \cmidrule(lr){1-22}
        \multicolumn{2}{c|}{ETTh2} &\textcolor{red}{\textbf{0.359}}  &\textcolor{red}{\textbf{0.391}}  &0.374 &0.398 &0.559 &0.515 &0.472  &0.465  &0.383  &0.408  &\textcolor{blue}{\textbf{0.365}} &\textcolor{blue}{\textbf{0.393}}  &0.383 &0.407 &0.387 &0.407 &0.526 &0.516 &0.414 &0.427 \\
        \cmidrule(lr){1-22}
        \multicolumn{2}{c|}{ECL} &\textcolor{red}{\textbf{0.171}}  &\textcolor{red}{\textbf{0.265}}  &0.219 &0.298 &0.212 &0.300 &0.189  &0.278  &0.217  &0.295  &\textcolor{blue}{\textbf{0.175}} &\textcolor{blue}{\textbf{0.269}} &0.178 &0.270 &0.216 &0.304 &0.193 &0.296 &0.192 &0.295 \\
        \cmidrule(lr){1-22}
        \multicolumn{2}{c|}{Exchange} &0.361  &\textcolor{blue}{\textbf{0.402}}  &0.378 &0.417 &0.354 &0.414 &\textcolor{red}{\textbf{0.329}}  &0.409  &\textcolor{blue}{\textbf{0.353}}  &\textcolor{red}{\textbf{0.399}} &0.374 &0.408 &0.360 &0.403 &0.367 &0.404 &0.461 &0.454 &0.416 &0.443 \\
        \cmidrule(lr){1-22}
        \multicolumn{2}{c|}{Traffic} &0.482  &0.315  &0.626 &0.378 &0.625 &0.383 &0.618  &0.390  &0.627  &0.376 &\textcolor{blue}{\textbf{0.431}} &\textcolor{blue}{\textbf{0.287}} &\textcolor{red}{\textbf{0.428}} &\textcolor{red}{\textbf{0.282}} &0.481 &0.304 &0.624 &0.340 &0.620 &0.336 \\
        \cmidrule(lr){1-22}
        \multicolumn{2}{c|}{Weather} &\textcolor{red}{\textbf{0.243}}  &\textcolor{red}{\textbf{0.271}}  &0.272 &0.291 &0.265 &0.317 &0.250  &0.270  &0.249  &0.276 &\textcolor{blue}{\textbf{0.246}} &\textcolor{blue}{\textbf{0.272}} &0.258 &0.278 &0.259 &0.281 &0.288 &0.314 &0.259 &0.287 \\
         \midrule[1pt]
        \multicolumn{2}{c|}{$1^{st}$ Count} &\textcolor{red}{\textbf{6}} &\textcolor{red}{\textbf{5}} &0 &0 &0 &0 &\textcolor{blue}{\textbf{1}} &0 &0 &\textcolor{blue}{\textbf{1}} &0 &\textcolor{blue}{\textbf{1}} &\textcolor{blue}{\textbf{1}} &\textcolor{blue}{\textbf{1}} &0 &0 &0 &0 &0 &0 \\
         \bottomrule[2pt]
    \end{tabular}
    }}
    \caption{Time series forecasting comparison. We set the lookback window size $L$ as 96 and the prediction length as $\tau \in \{96, 192, 336, 720\}$. The best results are in \textcolor{red}{red} and the second best are \textcolor{blue}{blue}. Results are averaged from all prediction lengths. Full results for all datasets are listed in Table \ref{tab:app_input_96} of Appendix \ref{results_appdenix}.}
    \label{tab:avg_input_96}
    \vspace{-5mm}
\end{table*}

\paragraph{SCI Block}
The data is composed of multiple channels of time series, usually following one common pattern, which we refer to as \textit{commonality}. Excluding this commonality, the remaining parts represent each channel's specific pattern, which we refer to as \textit{specificity}.

The commonality can be considered an abstract main channel, 
and it can be calculated as follow:
\begin{equation}\label{eq:commonality}
\begin{split}
    & X_{\text{Com}} = \operatorname{Compression_C}(X)
\end{split}
\end{equation}
where $\operatorname{Compression_C}: \mathbb{R}^{C}\mapsto \mathbb{R}^{1}$ contains two linear layers with intermediate LeakyReLU activation function and $X_{\text{Com}} \in \mathbb{R}^{1 \times L}$ is the commonality of $X \in \mathbb{R}^{C \times L}$. To further achieve the common temporal patterns, we employ a FFN and output the common pattern $X_{\text{Cp}} \in \mathbb{R}^{1\times L}$.


Then based on $X_{\text{Cp}}$ we can obtain the specificity temporal patterns, which can be formulated as:
\begin{equation}\label{eq:specificity}
\begin{split}
    X_{\text{Spc}} &= X - {X}_{\text{Cp}}, \\
    {X}_{\text{Sp}} &= \operatorname{FFN}(X_{\text{Spc}})
\end{split}
\end{equation}
where $X_{\text{Sp}} \in \mathbb{R}^{C \times L}$. Finally, we calculate the summation of $X_{\text{Cp}}$ and $X_{\text{Sp}}$ as the output $X_{\text{Sci}} \in \mathbb{R}^{C\times L}$ of SCI block.




\paragraph{Seasonal-Trend Forecaster}
Seasonal-trend decomposition as a basic function has been widely used in previous work~\cite{wu2021autoformer,zhou2022fedformer} for time series forecasting, and we also employ this in our model. Specifically, we first obtain the seasonal and trend components of $X_{\text{Sci}} \in \mathbb{R}^{C\times L}$ through the seasonal-trend decomposition as:
\begin{equation}
    X_{\text{Trend}}^{\text{Sci}}, X_{\text{Season}}^{\text{Sci}} = \operatorname{STD}({X}_{\text{Sci}}).
\end{equation}
Subsequently, we make predictions by applying two FFNs corresponding to the seasonal and trend components, which can be formulated as bellow:
\begin{equation}
\begin{split}
    Y_{\text{Trend}}^{\text{Sci}} &= \operatorname{Trend-FFN}(X_{\text{Trend}}^{\text{Sci}}), \\
    Y_{\text{Season}}^{\text{Sci}} &= \operatorname{Season-FFN}(X_{\text{Season}}^{\text{Sci}}), 
\end{split}
\end{equation}
where both $\operatorname{Trend-FFN(\cdot )}$ and $\operatorname{Season-FFN(\cdot )}$ contain two linear layers with intermediate LeakyReLU activation function. Then $Y_{\text{Trend}}^{\text{Sci}}$ and $Y_{\text{Season}}^{\text{Sci}}$ are concatenated as the output of the seasonal-trend forecaster block:
\begin{equation}
    Y = Y_{\text{Trend}}^{\text{Sci}} + Y_{\text{Season}}^{\text{Sci}}.
\end{equation}

\section{Experiments}

\subsection{Experiments Setup}

\paragraph{Datasets} We conduct extensive experiments on eight popular datasets, including  ETT datasets~\cite{zhou2021informer}, Electricity~\cite{wu2021autoformer}, Exchange~\cite{lai2018modeling}, Traffic~\cite{sen2019think} and Weather~\cite{wu2021autoformer}. More dataset details are in Appendix \ref{results_appdenix}.


\paragraph{Baselines} We select 10 highly regarded forecasting methods to serve as our benchmarks, including (i) Linear-based methods: RLinear~\cite{li2023revisiting}, DLinear~\cite{zeng2023transformers}, SparseTSF~\cite{lin2024sparsetsf}; (ii) Frequency-based methods: FreTS~\cite{yi2024frequency}, FITS~\cite{xu2023fits}; (iii) Transformer-based methods: Fredformer~\cite{piao2024fredformer}, iTransformer~\cite{liuitransformer}, PatchTST~\cite{nietime}, Stationary~\cite{liu2022non}; and (iv) TCN-based methods: TimesNet~\cite{wu2023timesnet}.

\paragraph{Implementation Details} All experiments in this study were carried out using PyTorch on one single NVIDIA RTX 3070 GPU with 8GB. We use Mean Squared Error (MSE) as the loss function and report the results using both MSE and Mean Absolute Error (MAE) as evaluation metrics.

\subsection{Main Results}

Table \ref{tab:avg_input_96} shows the average forecasting performance across four prediction lengths under a look-back window size of 96. Overall, our approach achieves leading performance on most datasets, securing 11 top-1 and 2 top-2 positions out of 16 in total across two metrics over eight datasets. 
We primarily attribute this to the energy amplification technique, which enhances the model's capability to model low-energy components.
However, we notice that the performance of Amplifier on the Traffic dataset is not outstanding. The reason is that for the Traffic dataset, which has strong periodicity, periodic information is typically represented by high-energy components. The energy amplification technique is designed to bring attention to neglected low-energy components, which may not be as beneficial for forecasting datasets with strong periodicity.
To further assess the performance of our model with varying lookback window sizes, we conduct additional experiments using a lookback window size of \(336\). The results, presented in Table \ref{tab:input_336}, indicate that as the lookback window size increases, Amplifier continues to demonstrate exceptional predictive performance.



\begin{figure*}[t!]
    \vspace{-2mm}
    \centering
    \subfigure[iTransformer]
    {
        \centering
        \includegraphics[width=0.235\linewidth]{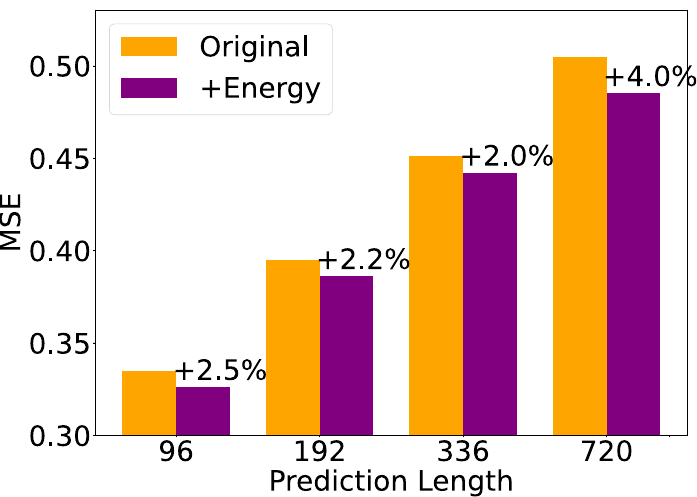}
    }
    \subfigure[Autoformer]
    {
        \centering
        \includegraphics[width=0.235\linewidth]{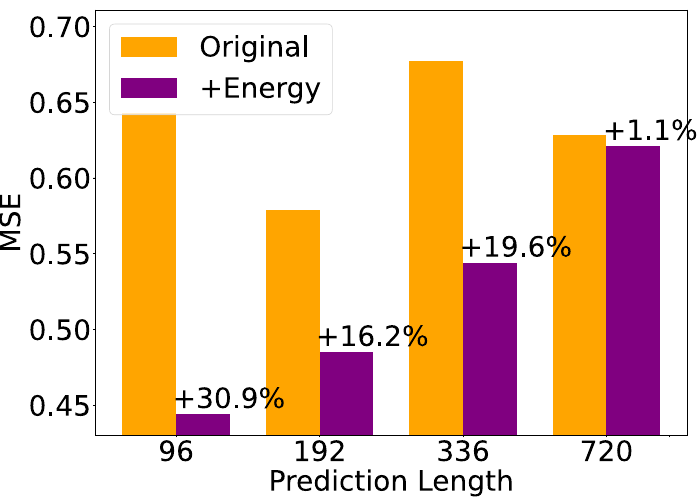}
    }
    \subfigure[SparseTSF]
    {
        \centering
        \includegraphics[width=0.235\linewidth]{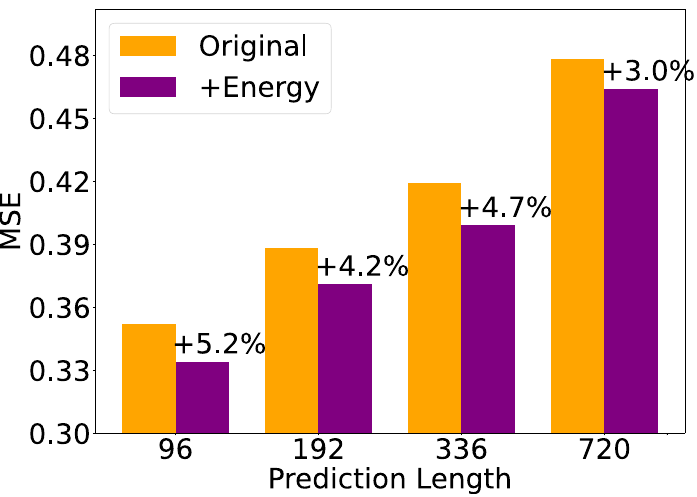}
    }
    \subfigure[DLinear]
    {
        \centering
        \includegraphics[width=0.235\linewidth]{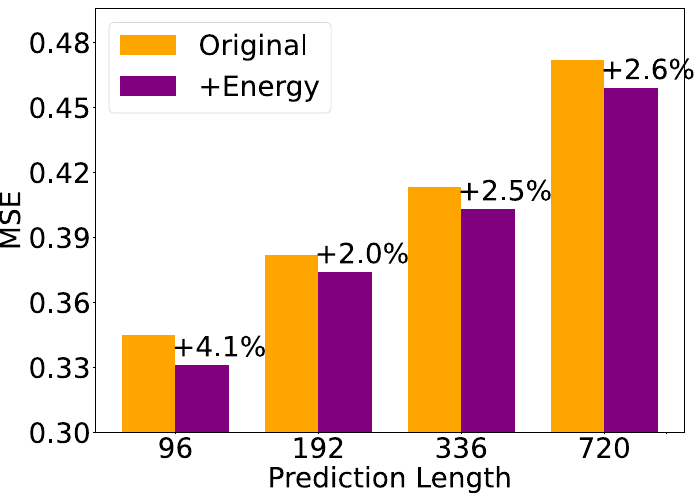}
    }
    \vspace{-4.5mm}
    \caption{Ablation prediction results of the energy amplification technique in four representative models.}
    \label{fig:Energy_Amplification_Boost}
    \vspace{-3mm}
\end{figure*}


\begin{table}[h]
    \centering
    {\fontsize{10pt}{10pt}\selectfont
    \scalebox{0.69}{
    \begin{tabular}{l l |c c | c c| c c}
    \toprule[2pt]
    \multicolumn{2}{c|}{Dataset} &\multicolumn{2}{c|}{ETTh1}  &\multicolumn{2}{c|}{ETTm2} &\multicolumn{2}{c}{Weather} \\ 
    \cmidrule(r){1-2}\cmidrule(lr){3-4}\cmidrule(lr){5-6}\cmidrule(lr){7-8}
       \multicolumn{2}{c|}{Metrics}  &MSE &MAE &MSE &MAE &MSE &MAE  \\
       \midrule[1pt]
       &w/o EAT    &0.393   &0.407   &0.182   &0.266    &0.184   &0.230   \\
       &Amplifier  &0.371  &0.392  &0.176 &0.258    &0.156 &0.205   \\  
       &\textcolor{red}{\textbf{Boost}} &\textcolor{red}{\textbf{5.643\%}}  &\textcolor{red}{\textbf{3.613\%}}   &\textcolor{red}{\textbf{3.484\%}}  &\textcolor{red}{\textbf{2.973\%}}    &\textcolor{red}{\textbf{15.100\%}}  &\textcolor{red}{\textbf{10.923\%}} \\   
    \bottomrule[2pt]
    \end{tabular}
    }}
    \caption{Ablation experiments of the energy amplification technique within Amplifier. w/o EAT refers to the version of Amplifier without the energy amplification technique. The ’Boost’ indicates the percentage of performance improvement after incorporating the energy amplification technique.}
    \label{tab:abl_energy_amplification}
    \vspace{-5mm}
\end{table}
\begin{table}[ht]
    \centering
    {\fontsize{10pt}{10pt}\selectfont
    \scalebox{0.59}{
    \begin{tabular}{l | c |c c|c c| c c |c c |c c}
    \toprule[2pt]
       \multicolumn{2}{c|}{Models}  &\multicolumn{2}{c|}{\textbf{Amplifier}}  &\multicolumn{2}{c|}{SparseTSF} &\multicolumn{2}{c|}{DLinear} &\multicolumn{2}{c|}{iTransformer} &\multicolumn{2}{c}{TimsNet}  \\
       \cmidrule(r){1-2}\cmidrule(lr){3-4}\cmidrule(lr){5-6}\cmidrule(lr){7-8}\cmidrule(lr){9-10}\cmidrule(lr){11-12}
       \multicolumn{2}{c|}{Metrics}  &MSE &MAE &MSE &MAE &MSE &MAE &MSE &MAE &MSE &MAE \\
       
       \midrule[1pt]
       \multirow{5}*{\rotatebox{90}{ETTm1}}
         &96   &\textcolor{red}{\textbf{0.288}}  &\textcolor{red}{\textbf{0.343}}   &0.306  &0.347 &\textcolor{blue}{\textbf{0.300}}  &\textcolor{blue}{\textbf{0.345}}  &0.303  &0.357  &0.335 &0.380   \\
         &192  &\textcolor{red}{\textbf{0.325}}   &\textcolor{red}{\textbf{0.366}}  &0.341  &\textcolor{blue}{\textbf{0.368}} &\textcolor{blue}{\textbf{0.336}}  &\textcolor{red}{\textbf{0.366}}    &0.345  &0.383  &0.358  &0.388   \\
         &336  &\textcolor{red}{\textbf{0.364}}   &\textcolor{blue}{\textbf{0.388}}  &0.373  &\textcolor{red}{\textbf{0.385}}  &\textcolor{blue}{\textbf{0.372}}  &0.390    &0.382  &0.405   &0.406  &0.418   \\
         &720  &0.430  &\textcolor{blue}{\textbf{0.423}}   &\textcolor{blue}{\textbf{0.429}}  &\textcolor{red}{\textbf{0.417}} &\textcolor{red}{\textbf{0.427}}  &\textcolor{blue}{\textbf{0.423}}    &0.443  &0.439  &0.449  &0.443   \\
         \cmidrule(lr){2-12}
         &Avg &\textcolor{red}{\textbf{0.352}} &\textcolor{blue}{\textbf{0.380}} &0.362  &\textcolor{red}{\textbf{0.379}}  &\textcolor{blue}{\textbf{0.359}}  &0.381 &0.368  &0.396  &0.387   &0.407    \\
         \midrule[1pt]
         
         \multirow{5}*{\rotatebox{90}{ETTh1}}
         &96   &\textcolor{red}{\textbf{0.371}}  &\textcolor{red}{\textbf{0.395}}  &0.393  &0.407  &\textcolor{blue}{\textbf{0.384}}  &\textcolor{blue}{\textbf{0.405}} &0.402  &0.418  &0.398  &0.418   \\
         &192  &\textcolor{red}{\textbf{0.408}}  &\textcolor{red}{\textbf{0.415}}   &\textcolor{blue}{\textbf{0.421}}  &\textcolor{blue}{\textbf{0.423}}  &0.430  &0.442 &0.450  &0.449 &0.447  &0.449   \\
         &336  &\textcolor{red}{\textbf{0.396}}  &\textcolor{red}{\textbf{0.416}} &\textcolor{blue}{\textbf{0.401}}  &\textcolor{blue}{\textbf{0.425}}  &0.447  &0.448 &0.479  &0.470 &0.493  &0.468 \\
         &720  &\textcolor{blue}{\textbf{0.451}}  &\textcolor{blue}{\textbf{0.463}} &\textcolor{red}{\textbf{0.421}}  &\textcolor{red}{\textbf{0.446}}  &0.504  &0.515 &0.584  &0.548 &0.518  &0.504   \\
         \cmidrule(lr){2-12}
         &Avg &\textcolor{red}{\textbf{0.407}}  &\textcolor{red}{\textbf{0.422}}  &\textcolor{blue}{\textbf{0.409}}  &\textcolor{blue}{\textbf{0.425}}  &0.441  &0.453 &0.479  &0.471 &0.464   &0.460    \\
         \midrule[1pt]

         \multirow{5}*{\rotatebox{90}{Weather}} 
         &96   &\textcolor{red}{\textbf{0.150}}  &\textcolor{red}{\textbf{0.203}}  &0.177  &0.227  &0.175  &0.235  &\textcolor{blue}{\textbf{0.164}}  &\textcolor{blue}{\textbf{0.216}}  &0.172  &0.220  \\
         &192  &\textcolor{red}{\textbf{0.192}}  &\textcolor{red}{\textbf{0.242}}  &0.221  &0.264  &0.218  &0.278 &\textcolor{blue}{\textbf{0.205}}  &\textcolor{blue}{\textbf{0.251}} &0.219  &0.261  \\
         &336  &\textcolor{red}{\textbf{0.241}}  &\textcolor{red}{\textbf{0.280}}  &0.267  &0.296   &0.263  &0.314 &\textcolor{blue}{\textbf{0.256}}  &\textcolor{blue}{\textbf{0.290}} &0.280  &0.306  \\
         &720  &\textcolor{red}{\textbf{0.316}}  &\textcolor{red}{\textbf{0.332}}  &0.334  &0.343  &\textcolor{blue}{\textbf{0.324}}  &0.362 &0.326  &\textcolor{blue}{\textbf{0.338}} &0.365  &0.359   \\
         \cmidrule(lr){2-12}
         &Avg &\textcolor{red}{\textbf{0.225}}  &\textcolor{red}{\textbf{0.264}}  &0.250  &0.283  &0.245  &0.297 &\textcolor{blue}{\textbf{0.238}}  &\textcolor{blue}{\textbf{0.274}} & 0.259  &0.287   \\
         
         \midrule[1pt]
         
         \multirow{5}*{\rotatebox{90}{Electricity}} 
         &96   &\textcolor{red}{\textbf{0.133}}  &\textcolor{blue}{\textbf{0.231}}  &0.147  &0.240  &\textcolor{blue}{\textbf{0.140}}  &0.237 &\textcolor{red}{\textbf{0.133}}  &\textcolor{red}{\textbf{0.229}}  &0.168  &0.272 \\
         &192  &\textcolor{blue}{\textbf{0.156}}  &0.252  &0.158  &\textcolor{blue}{\textbf{0.251}}   &\textcolor{red}{\textbf{0.154}}  &\textcolor{red}{\textbf{0.250}} &\textcolor{blue}{\textbf{0.156}}  &\textcolor{blue}{\textbf{0.251}} &0.184  &0.289  \\
         &336  &\textcolor{red}{\textbf{0.169}}  &\textcolor{red}{\textbf{0.266}}  &0.174  &0.268  &\textcolor{red}{\textbf{0.169}}  &0.268 &\textcolor{blue}{\textbf{0.172}}  &\textcolor{blue}{\textbf{0.267}} &0.198  &0.300  \\
         &720  &\textcolor{red}{\textbf{0.199}}  &\textcolor{red}{\textbf{0.296}}  &0.212  &\textcolor{blue}{\textbf{0.299}}  &\textcolor{blue}{\textbf{0.204}} &0.300 &0.209  &0.304  &0.220  &0.320   \\
         \cmidrule(lr){2-12}
         &Avg &\textcolor{red}{\textbf{0.164}}  &\textcolor{red}{\textbf{0.261}}  &0.173  &0.265  &\textcolor{blue}{\textbf{0.167}}  &0.264 &0.168  &\textcolor{blue}{\textbf{0.263}}  &0.193   &0.287   \\
        \midrule[1pt]
         \multicolumn{2}{c|}{$1^{st}$ Count} &\textcolor{red}{\textbf{17}}  &\textcolor{red}{\textbf{14}}  &1  &\textcolor{blue}{\textbf{4}}  &\textcolor{blue}{\textbf{3}}  &2  &1  &1  &0  &0  \\
         
         \bottomrule[2pt]
    \end{tabular}
    }}
    \caption{Time series forecasting comparison with the lookback window size $L$ as 336 and the prediction length as $\tau \in \{96, 192, 336, 720\}$. The best results are in \textcolor{red}{red} and the second best are in \textcolor{blue}{blue}. Avg means the average results from all four prediction lengths.}
    \label{tab:input_336}
    \vspace{-6mm}
\end{table}

\subsection{Model Analysis}

\paragraph{Effectiveness Analysis of the Energy Amplification Technique} In this part, we investigate the effectiveness of the energy amplification technique from two perspectives: the role it plays within Amplifier and the performance improvements it brings when integrated as a general technique into other foundational models.
As shown in Table \ref{tab:abl_energy_amplification}, we compare Amplifier and w/o EAT which is the version of Amplifier without the energy amplification technique (EAT), on the ETTh1, ETTm2, and Weather datasets. It demonstrates that the energy amplification technique can improve predictive performance by increasing the model's attention to low-energy components. It's worth noting that, on the weather dataset, the impact of using the energy amplification technique on prediction results is significant, affecting MSE and MAE by as much as 15.100\% and 10.923\%, respectively. As illustrated by the butterfly effect~\cite{lorenz1972predictability}, small changes in a weather model (such as a butterfly flapping its wings in Brazil) can trigger large-scale atmospheric changes far away (such as a hurricane in the United States) through a series of causal relationships. Effectively capturing and modeling these low-energy components can significantly improve the accuracy of weather predictions.


    
    

Besides, the energy amplification technique can not only function as a built-in module within Amplifier but also as a universal technology that can be integrated into other forecasting models, such as Transformer-based models: iTransformer and Autoformer~\cite{wu2021autoformer}; MLP-based models: SparseTSF~\cite{lin2024sparsetsf} and DLinear~\cite{zeng2023transformers}. As illustrated in Figure \ref{fig:Energy_Amplification_Boost}, on the ETTm1 dataset, Transformer-based models achieved improvements of 9.837\% in MSE and 4.236\% in MAE, while Linear-based models achieved improvements of 3.603\% in MSE and 2.244\% in MAE (the statistical values are in the sense of average). These results clearly demonstrate the effectiveness of the energy amplification technique, showing that it can significantly enhance the performance of foundational models in time series forecasting.




\paragraph{Effectiveness Analysis of SCI Block} The SCI block enhances temporal relationship modeling by improving information utilization, specifically through the consideration of interactions between channels. We conduct ablation experiments on the role of the SCI block within Amplifier, as shown in Table \ref{tab:abl_SCI}, where w/o SCI refers to the version without the SCI block. We choose the Electricity and Traffic datasets with a huge number of channels for our experiments because the interactions between channels becomes more pronounced as the number of channels increases. Specifically, using the SCI block achieved improvements of 7.479\% in MSE and 3.007\% in MAE on the Electricity dataset, while showed improvements of 5.018\% in MSE and 4.294\% in MAE on the Traffic dataset. Therefore, the SCI block contributes to enhancing model performance by effectively leveraging interactions between channels.
\vspace{-1mm}

\begin{table}[!t]
    \centering
    {\fontsize{10pt}{10pt}\selectfont
    \scalebox{0.53}{
    \begin{tabular}{c |c |c c| c c| c c |c c }
    \toprule[2pt]
       \multicolumn{2}{c|}{Horizon}  &\multicolumn{2}{c|}{96}  &\multicolumn{2}{c|}{192} &\multicolumn{2}{c|}{336} &\multicolumn{2}{c}{720}  \\
       \cmidrule(r){1-2}\cmidrule(lr){3-4}\cmidrule(lr){5-6} \cmidrule(lr){7-8} \cmidrule(lr){9-10}
       \multicolumn{2}{c|}{Metrics}&MSE &MAE  &MSE &MAE &MSE &MAE &MSE &MAE  \\
       \midrule[1pt] 
       \multirow{3}*{\rotatebox{90}{ECL}}
       &w/o SCI &0.156  &0.247  &0.167  &0.256  &0.189  &0.278  &0.230  &0.312 \\
        &Amplifier  &0.147  &0.243  &0.157  &0.251  &0.174   &0.269   &0.206   &0.296 \\
        &\textcolor{red}{\textbf{Boost}} &\textcolor{red}{\textbf{5.769\%}}  &\textcolor{red}{\textbf{1.619\%}}  &\textcolor{red}{\textbf{5.988\%}}  &\textcolor{red}{\textbf{1.953\%}}  &\textcolor{red}{\textbf{7.839\%}}  &\textcolor{red}{\textbf{3.237\%}}  &\textcolor{red}{\textbf{10.318\%}}  &\textcolor{red}{\textbf{5.219\%}} \\

        \midrule[1pt] 
       \multirow{3}*{\rotatebox{90}{Traffic}}
       &w/o SCI &0.500  &0.328  &0.477  &0.305  &0.509  &0.326  &0.544  &0.345 \\
        &Amplifier  &0.455   &0.298   &0.470   &0.304   &0.479   &0.316   &0.523   &0.328  \\
        &\textcolor{red}{\textbf{Boost}} &\textcolor{red}{\textbf{9.000\%}}  &\textcolor{red}{\textbf{9.063\%}}  &\textcolor{red}{\textbf{1.426\%}}  &\textcolor{red}{\textbf{0.230\%}}  &\textcolor{red}{\textbf{5.838\%}}  &\textcolor{red}{\textbf{3.067\%}}  &\textcolor{red}{\textbf{3.807\%}}  &\textcolor{red}{\textbf{4.817\%}} \\

         \bottomrule[2pt]
    \end{tabular}
    }}
    \caption{Ablation experiments for SCI block of Amplifier where w/o SCI refers to the version without the SCI block. The 'Boost' indicates the percentage of performance improvement after equipping with the SCI block.}
    \label{tab:abl_SCI}
    \vspace{-4mm}
\end{table}

\begin{figure}
    \centering
    \includegraphics[width=1\linewidth]{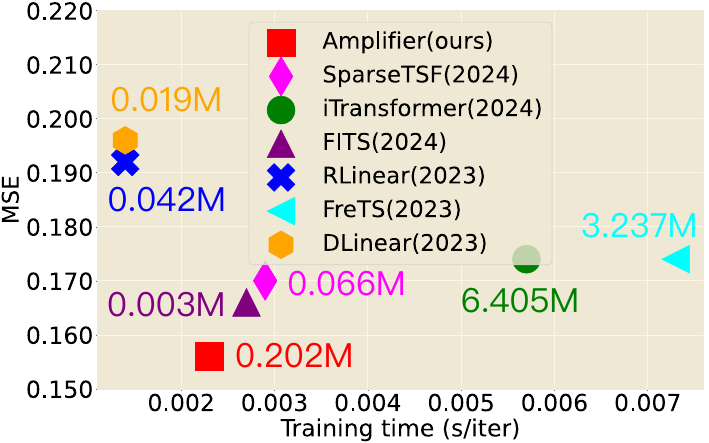}
    \vspace{-4mm}
    \caption{Model efficiency comparison in terms of MSE, the scale of parameters, and training speed.}
    \label{fig:parameter}
    \vspace{-4mm}
\end{figure}


\subsection{Efficiency Analysis}
The theoretical complexity of the Amplifier is $\mathcal{O}(L \log L)$.
We conduct a comprehensive comparison of the forecasting performance, the scale of parameters, and the training speed of the following representative models, including SparseTSF, RLinear, DLinear, FITS, FreTS, and iTransformer. 
We choose the Weather dataset under the scenario of $L=96$ and $\tau=96$ for comparison. 


From Figure \ref{fig:parameter}, it clear to see that the scale of parameters and training speed of Amplifier are at a medium level. However, it's noteworthy that both SparseTSF and FITS consider lightweight architecture as one of their main contributions, whereas Amplifier does not have this as a primary goal in its model design. As for RLinear and DLinear, since Amplifier aims to enhance the model’s attention on low-energy components while avoiding any negative impact on modeling high-energy components, it requires a dedicated model component to handle low-energy components. This inevitably results in the Amplifier having a larger parameter size and longer training time compared to these two models. 





\begin{figure}[h!]
    \centering
    \vspace{-2mm}
    \includegraphics[width=1\linewidth]{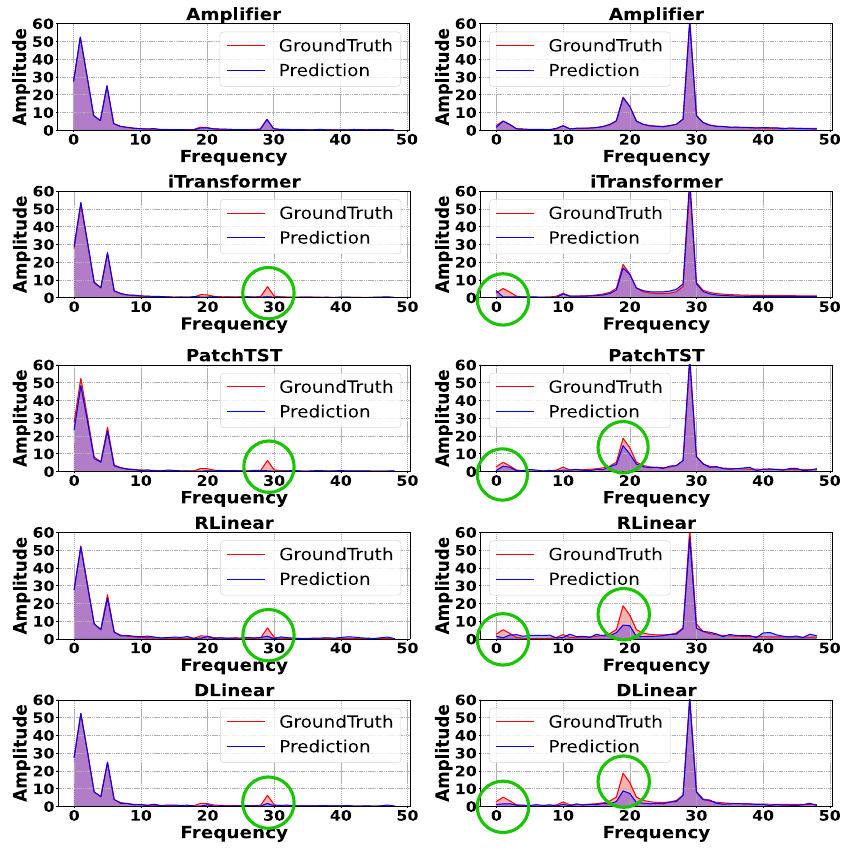}
    \caption{Prediction spectrums of both Amplifier and four SOTA models on two synthetic signals. The small green circles represent the neglected low-energy components.}
    \label{fig:motivation_full}
    \vspace{-3mm}
\end{figure}

\subsection{Visualizations}
\paragraph{Visualization Analysis about the  Energy Amplification Technique} To intuitively validate the modeling capability of our amplification technique in low energy components, we conduct experiments on two simple synthetic signals, and compare our model with the four representative models, including iTransformer, PatchTST, RLinear, and DLinear.
The results are shown in Figure \ref{fig:motivation_full}, which demonstrates that Amplifier can handle low-energy components as effectively as high-energy components. Compared with our model, the four models neither can fully model the low-energy components, regardless of whether these low-energy components appear at high-frequency region or low-frequency region.

\begin{figure}[h!]
    \centering
    \vspace{-2mm}
    \includegraphics[width=1\linewidth]{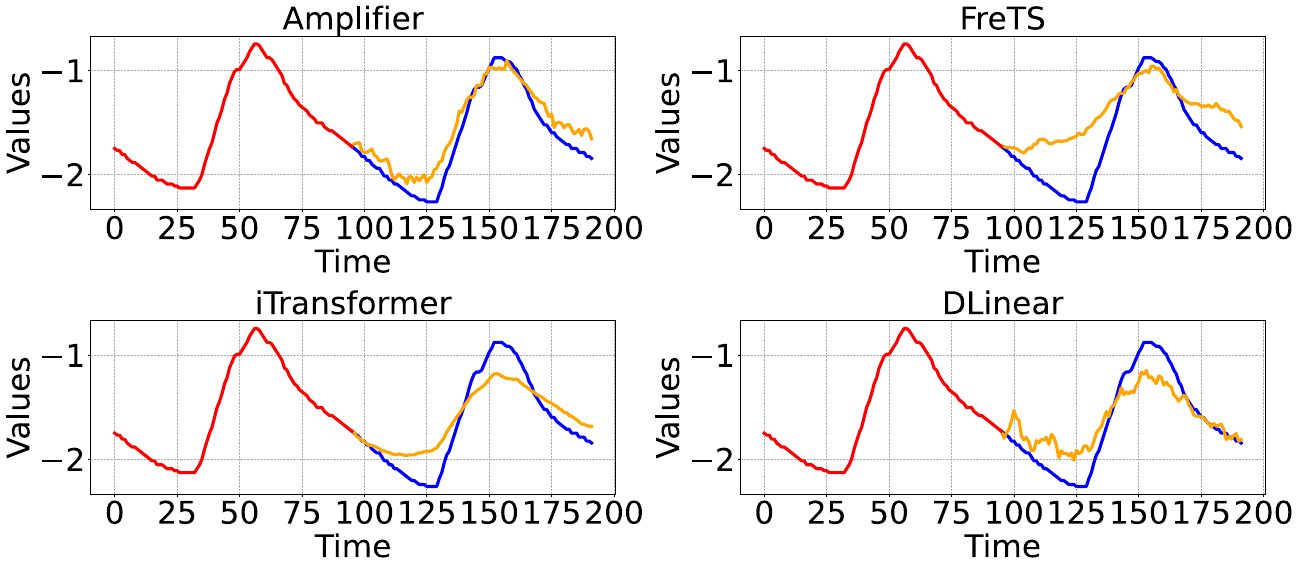}
        \caption{Visualization of prediction results on the ETTm2 dataset ($96 \to 96$).}
    \label{fig:compare_ETTm2}
    \vspace{-4mm}
\end{figure}

\begin{figure}[h!]
    \centering
    \includegraphics[width=1\linewidth]{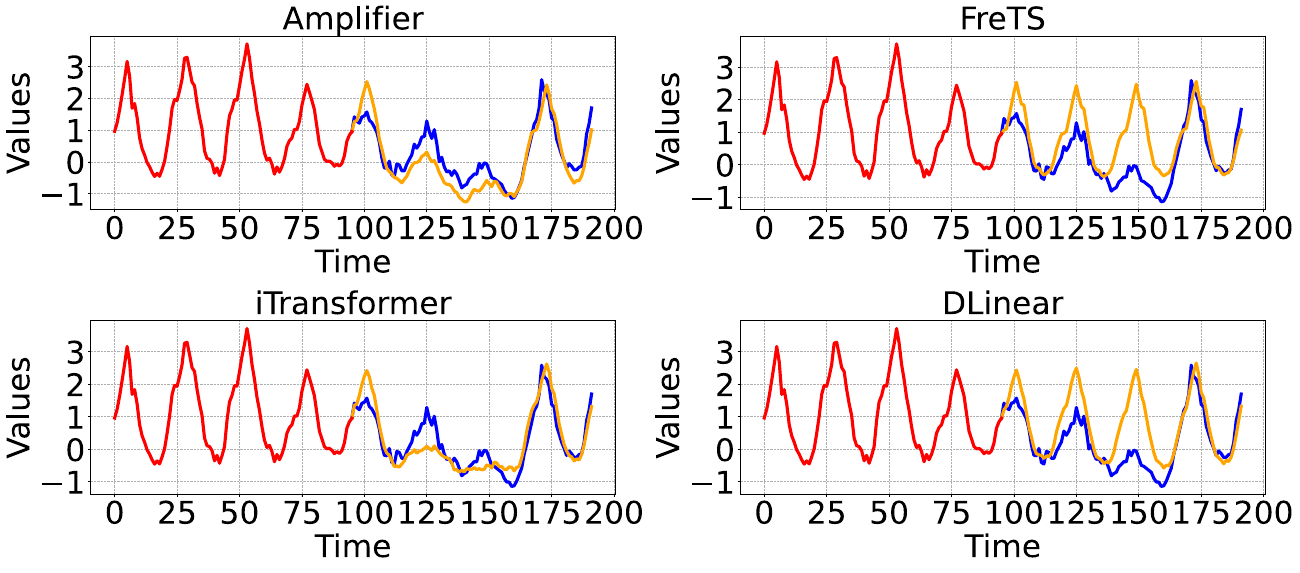}
    \vspace{-3mm}
    \caption{Visualization of prediction results on the Electricity dataset ($96 \to 96$).}
    \label{fig:compare_ECL}
    \vspace{-3mm}
\end{figure}

\paragraph{Visualization of Forecasting Results} To visually compare the performance of Amplifier with state-of-the-art models, including iTransformer, FreTS, and DLinear, we present prediction showcases on the ETTm2 and Electricity datasets, and the results are shown in Figures \ref{fig:compare_ETTm2} and \ref{fig:compare_ECL}. The red line represents the input sequence, the blue line represents the ground truth, and the yellow line represents the predicted value.
Compared to these different types of state-of-the-art models, Amplifier provides the most accurate predictions of future series variations, demonstrating superior performance.

\section{Conclusion}

Considering that low-energy components is often overlooked in existing methods, we propose the energy amplification technique composed of the energy amplification block and energy restoration block. The core idea of this technique is to increase the model's attention to low-energy components by flipping spectrum to amplify the energy of those components, thereby enhancing the model's ability to process low-energy components. To better leverage the energy amplification technique, we design a novel model called Amplifier for time series forecasting. Comprehensive empirical experiments on eight real-world datasets have validated the superiority of our proposed model.

\section*{Acknowledgments}
This research was funded by the National Natural Science Foundation of China, Grant No. 62272048.

\bibliography{aaai25}

\clearpage
\appendix

\section{Theoretical Proofs}\label{proof_appendix}
\setcounter{myTheo}{0}

To clarify the proof process, we first define the frequency domain representation of a time domain signal and introduce a lemma to assist the proof process.
\begin{myDef}\label{def_1}
In the time domain, a signal $X(t)$ is a function of time $t$ and can be represented as a sum of various frequency components~\cite{hayes1996statistical}, each with a specific amplitude and phase:
\begin{equation}
\begin{split}
    X(t) &= \sum_{f}\mathcal{X}[f], \\
    \mathcal{X}[f] &= \mathcal{A}_{f} e^{j\phi_{f}} = \mathcal{A}_{f} e^{j\left(\omega_{f} t+\psi_{f}\right)},
\end{split}
\end{equation}
where $\phi_{f}=\omega_{f}t+\psi_{f}$ refers to the total phase, $\mathcal{A}_{f}$, $\omega_{f}(\omega_{f}=2\pi f)$ and $\psi_{f}$ represent the amplitude, angular frequency, and initial phase of a particular frequency component $f$, respectively.
\end{myDef}

\begin{myLemma}\label{lemma_1}
Given that $c=e^{j\theta_{1}}$ and $d=e^{j\theta_{2}}$ are complex numbers with $\|c\|=\|d\|=1$, then $\|a c-b d\|^{2}$ can be expanded as follows:
\begin{equation}
    \|a c-b d\|^{2}=\|a\|^{2}+\|b\|^{2}-2ab\cos \left(\theta_{1}-\theta_{2}\right).
\end{equation}
\end{myLemma}

\noindent
\textbf{Proof of Lemma 1}
\begin{align}
    \|a c-b d\|^{2} &= (a c-b d)(\overline{a c}-\overline{b d}) \notag \\
    &=a c \overline{a c}-a c \overline{b d}-b d \overline{a c}+b d \overline{b d} \notag \\
    &=\|a\|^{2}\|c\|^{2}-acb\bar{d}-bda \bar{c}+\|b\|^{2}\|d\|^{2} \notag \\
    &=\|a\|^{2}+\|b\|^{2}-ab(c \bar{d}+\bar{c}d) \notag \\
    &=\|a\|^{2}+\|b\|^{2}-2ab\cos \left(\theta_{1}-\theta_{2}\right) \notag
\end{align}
Proved.

\begin{myTheo}\label{theorem_1}
In the initial stage of network training, the  loss of high-energy components $\mathcal{L}(\mathcal{Y}_{H},\mathcal{\hat{Y}}_{H};\Theta_{H})$ occupies a significantly larger proportion of the overall loss compared to the loss of low-energy components $\mathcal{L}(\mathcal{Y}_{L},\mathcal{\hat{Y}}_{L};\Theta_{L})$, that is:
\begin{equation}
    \frac{\mathcal{L}(\mathcal{Y}_{H},\hat{\mathcal{Y}}_{H};\Theta_{H})}{\mathcal{L}(\mathcal{Y},\mathcal{\hat{Y}};\Theta)} \gg \frac{\mathcal{L}(\mathcal{Y}_{L},\mathcal{\hat{Y}}_{L};\Theta_{L})}{\mathcal{L}(\mathcal{Y},\mathcal{\hat{Y}};\Theta)}.
\end{equation}
\end{myTheo}
\noindent
\textbf{Proof of Theorom 1}

As we have mentioned in Preliminaries section, since the data can be divided into high-energy components and low-energy components, the overall loss can be composed of the individual losses of these two components as below:
\begin{equation}\label{eq:loss}
    \mathcal{L}(\mathcal{Y},\hat{\mathcal{Y}};\Theta) = \mathcal{L}(\mathcal{Y}_{H},\hat{\mathcal{Y}}_{H};\Theta_{H}) + \mathcal{L}(\mathcal{Y}_{L},\hat{\mathcal{Y}}_{L};\Theta_{L}).
\end{equation}
Combine Eq (\ref{eq:loss}) and Definition 1\label{def_1}, $\mathcal{L}(\mathcal{Y}_{H},\hat{\mathcal{Y}}_{H};\Theta_{H})$ and $\mathcal{L}(\mathcal{Y}_{L},\hat{\mathcal{Y}}_{L};\Theta_{L})$ can be rewritten separately as follows:
\begin{align}\label{eq:Loss_H}
    & \mathcal{L}(\mathcal{Y}_{H},\hat{\mathcal{Y}}_{H};\Theta_{H}) \notag \\
    & =  \left \| \mathcal{Y}_{H} - \hat{\mathcal{Y}}_{H}\right \| ^{2}_{2} \notag \\
    & = \left \| \mathcal{A}_{H} e^{j\phi_{H}} - \hat{\mathcal{A}}_{H} e^{j\hat{\phi}_{H}}\right \| ^{2}_{2},
\end{align}
where the subscript $H$ is an abbreviation for high-energy components. $\mathcal{A}_{H}$ and $\phi_{H}$ represent the true values of the amplitude and the total phase of the high-energy components, while $\Theta_{H} = \left \{ \hat{\mathcal{A}}_{H}, \hat{\phi}_{H} \right \}$ refers to the corresponding values learned by the neural network.

\begin{align}\label{eq:Loss_L}
    & \mathcal{L}(\mathcal{Y}_{L},\hat{\mathcal{Y}}_{L};\Theta_{L}) \notag \\
    & =  \left \| \mathcal{Y}_{L} - \hat{\mathcal{Y}}_{L}\right \| ^{2}_{2} \notag \\
    & = \left \| \mathcal{A}_{L} e^{j\phi_{L}} - \hat{\mathcal{A}}_{L} e^{j\hat{\phi}_{L}}\right \| ^{2}_{2},
\end{align}
where the subscript $L$ is an abbreviation for low-energy components. $\mathcal{A}_{L}$ and $\phi_{L}$ represent the true values of the amplitude and the total phase of the low-energy components, while $\Theta_{L} = \left \{ \hat{\mathcal{A}}_{L}, \hat{\phi}_{L}\right \}$ refers to the corresponding values learned by the neural network.


Based on Lemma 1, Eq (\ref{eq:Loss_H}) and Eq (\ref{eq:Loss_L}) can be expanded as:
\begin{align}\label{eq:Loss_H_exp}
    & \mathcal{L}(\mathcal{Y}_{H},\hat{\mathcal{Y}}_{H};\Theta_{H}) \notag \\
    & = \left \| \mathcal{A}_{H} e^{j\phi_{H}} - \hat{\mathcal{A}}_{H} e^{j\hat{\phi}_{H}} \right \| ^{2}_{2} \notag \\
    & = \left \| \mathcal{A}_{H} \right \|^{2}_{2} + \left \| \hat{\mathcal{A}}_{H} \right \|^{2}_{2} -\mathcal{A}_{H}\hat{\mathcal{A}}_{H}(e^{j\phi_{H}}e^{-j\hat{\phi}_{H}} + e^{-j\phi_{H}}e^{j\hat{\phi}_{H}}) \notag \\
    & = \left\| \mathcal{A}_{H}\right\|_{2}^{2}+\left\|\hat{\mathcal{A}}_{H}\right\|_{2}^{2}-\mathcal{A}_{H} \hat{\mathcal{A}}_{H} \Gamma_{H},
\end{align}
where $\Gamma_{H} = 2 \cos \left(\phi_{H}-\hat{\phi}_{H}\right)$.
\begin{align}\label{eq:Loss_L_exp}
    & \mathcal{L}(\mathcal{Y}_{L},\hat{\mathcal{Y}}_{L};\Theta_{L}) \notag \\
    & = \left \| \mathcal{A}_{L} e^{j\phi_{L}} - \hat{\mathcal{A}}_{L} e^{j\hat{\phi}_{L}} \right \| ^{2}_{2} \notag \\
    & = \left \| \mathcal{A}_{L} \right \|^{2}_{2} + \left \| \hat{\mathcal{A}}_{L} \right \|^{2}_{2} -\mathcal{A}_{L}\hat{\mathcal{A}}_{L}(e^{j\phi_{L}}e^{-j\hat{\phi}_{L}} + e^{-j\phi_{L}}e^{j\hat{\phi}_{L}}) \notag \\
    & = \left\|\mathcal{A}_{L}\right\|_{2}^{2}+\left\|\hat{\mathcal{A}}_{L}\right\|_{2}^{2}-\mathcal{A}_{L} \hat{\mathcal{A}}_{L} \Gamma_{L},
\end{align}
where $\Gamma_{L} = 2 \cos \left(\phi_{L}-\hat{\phi}_{L}\right)$.



In the initial stages of neural network training, the parameters are randomly initialized and their values are typically small~\cite{glorot2010understanding, Goodfellow-et-al-2016}. So based on Eq (\ref{eq:Loss_H_exp}) and Eq (\ref{eq:Loss_L_exp}), $\mathcal{L}(\mathcal{Y}_{H},\hat{\mathcal{Y}}_{H};\Theta_{H})$ and $\mathcal{L}(\mathcal{Y}_{L},\hat{\mathcal{Y}}_{L};\Theta_{L})$ can be approximately expressed in the following forms:
\begin{equation}\label{eq:loss_h_approx}
    \mathcal{L}(\mathcal{Y}_{H},\hat{\mathcal{Y}}_{H};\Theta_{H})\approx  \left\|\mathcal{A}_{H}\right\|_{2}^{2},
\end{equation}
\begin{equation}\label{eq:loss_l_approx}
    \mathcal{L}(\mathcal{Y}_{L},\hat{\mathcal{Y}}_{L};\Theta_{L})\approx  \left\|\mathcal{A}_{L}\right\|_{2}^{2}.
\end{equation}
Since the amplitude of the high-energy components is much greater than that of the low-energy components, i.e. $\left \| \mathcal{A}_{H} \right \| \gg \left \| \mathcal{A}_{L}  \right \| $, then $\left \| \mathcal{A}_{H} \right \|^{2}_{2} \gg \left \| \mathcal{A}_{L}  \right \| ^{2}_{2}$. Combining Eq (\ref{eq:loss_h_approx}) and Eq (\ref{eq:loss_l_approx}), we obtain:
\begin{equation}
    \mathcal{L}\left(\mathcal{Y}_{H}, \hat{\mathcal{Y}}_{H} ; \Theta_{H}\right)\gg \mathcal{L}\left(\mathcal{Y}_{L}, \hat{\mathcal{Y}}_{L} ; \Theta_{L}\right),
\end{equation}
that is:
\begin{equation}
    \frac{\mathcal{L}(\mathcal{Y}_{H},\hat{\mathcal{Y}}_{H};\Theta_{H})}{\mathcal{L}(\mathcal{Y},\hat{\mathcal{Y}};\Theta)} \gg \frac{\mathcal{L}(\mathcal{Y}_{L},\hat{\mathcal{Y}}_{L};\Theta_{L})}{\mathcal{L}(\mathcal{Y},\hat{\mathcal{Y}};\Theta)}.
\end{equation}
Proved.

\begin{myTheo}\label{theorem_2}
Parameter updates are influenced by the energy of their corresponding components, meaning that the updates for parameters $\Theta_L$ related to low-energy components are much less efficient than those $\Theta_H$ for high-energy components, which can be expressed as:
\begin{equation}
\frac{\mathcal{L}(\mathcal{Y}_{H},\mathcal{\hat{Y}}_{H};\Theta_{H})}{\partial \Theta_H} \gg \frac{\mathcal{L}(\mathcal{Y}_{L},\mathcal{\hat{Y}}_{L};\Theta_{L})}{\partial \Theta_L}.
\end{equation}
\end{myTheo}

\noindent
\textbf{Proof of Theorom 2}

Firstly, according to Definition 1, $\mathcal{L}(\mathcal{Y}_{H},\hat{\mathcal{Y}}_{H};\Theta_{H})$ and $\mathcal{L}(\mathcal{Y}_{L},\hat{\mathcal{Y}}_{L};\Theta_{L})$ can be expanded in detail as follows:
\begin{align}
    & \mathcal{L}(\mathcal{Y}_{H},\hat{\mathcal{Y}}_{H};\Theta_{H}) \notag \\
    &=  \left \| \mathcal{Y}_{H} - \hat{\mathcal{Y}}_{H}\right \| ^{2}_{2} \notag \\
    & = \left \| \mathcal{A}_{H} e^{j\left(\omega_{H}t+\psi_{H}\right)} - \hat{\mathcal{A}}_{H} e^{j\left(\hat{\omega}_{H}t+\hat{\psi}_{H}\right)}\right \| ^{2}_{2},
\end{align}
\begin{align}
    & \mathcal{L}(\mathcal{Y}_{L},\hat{\mathcal{Y}}_{L};\Theta_{L}) \notag \\
    &=  \left \| \mathcal{Y}_{L} - \hat{\mathcal{Y}}_{L}\right \| ^{2}_{2} \notag \\
    & = \left \| \mathcal{A}_{L} e^{j\left(\omega_{L}t+\psi_{L}\right)} - \hat{\mathcal{A}}_{L} e^{j\left(\hat{\omega}_{L}t+\hat{\psi}_{L}\right)}\right \| ^{2}_{2},
\end{align}
where $\hat{\mathcal{A}}_{H}$, $\hat{\omega}_{H}$, and $\hat{\psi}_{H}$ refer to the parameters related to the high-energy components while $\hat{\mathcal{A}}_{L}$, $\hat{\omega}_{L}$, and $\hat{\psi}_{L}$ represent the parameters associated with the low-energy components. 

Then based on the common time-domain parameter update algorithms~\cite{bottou2010large}, we derive the parameter update algorithms in the frequency domain:
\begin{align}
    \hat{\mathcal{A}}^{i+1}_{f} &= \hat{\mathcal{A}}^{i}_{f}-\eta \frac{\partial \mathcal{L}}{\partial \hat{\mathcal{A}}^{i}_{f}} \notag \\
                &=\hat{\mathcal{A}}^{i}_{f} - \eta e^{j\left(\hat{\omega}^{i}_{f} t+\hat{\psi}^{i}_{f}\right)}, \label{eq_amplitude} \\
    \hat{\omega }^{i+1}_{f} &= \hat{\omega }^{i}_{f}-\eta \frac{\partial \mathcal{L}}{\partial \hat{\omega}^{i}_{f}} \notag \\
                &=\hat{\omega }^{i}_{f} - \eta \hat{\mathcal{A}}^{i}_{f}te^{j\left(\hat{\omega}^{i}_{f} t+\hat{\psi}^{i}_{f}\right)}, \label{eq_omega}  \\
    \hat{\psi }^{i+1}_{f} &= \hat{\psi }^{i}_{f}-\eta \frac{\partial \mathcal{L}}{\partial \hat{\psi }^{i}_{f}} \notag \\
                &=\hat{\psi }^{i}_{f} - \eta \hat{\mathcal{A}}^{i}_{f}e^{j\left(\hat{\omega}^{i}_{f} t+\hat{\psi}^{i}_{f}\right)}, \label{eq_psi}
\end{align}
where subscript $f$ denotes the value at frequency point $f$, superscript $i$ represents the $i$-th update iteration, and $\eta$ refers to the learning rate. 
Since the amplitude $\hat{\mathcal{A}}_{L}$ of low-energy components is much smaller than that of high-energy components, the parameters $\hat{\omega}_{L}$ and $\hat{\psi}_{L}$ related to low-energy components are updated to a much lesser extent each iteration compared to those related to high-energy components, according to Eq (\ref{eq_omega}) and Eq (\ref{eq_psi}). At the same time, the inefficient updates of parameters $\hat{\omega}_{L}$ and $\hat{\psi}_{L}$ also hinder the effective update of parameter $\hat{\mathcal{A}}_{L}$ in Eq (\ref{eq_amplitude}). 

In summary, the update efficiency of parameters $ \Theta_{L} = \{ \hat{\mathcal{A}}_{L}, \hat{\omega}_{L}, \hat{\psi}_{L} \}$ related to low-energy components is significantly lower than that of the parameters related to high-energy components, i.e.,
\begin{equation}
\frac{\mathcal{L}(\mathcal{Y}_{H},\hat{\mathcal{Y}}_{H};\Theta_{H})}{\partial \Theta_H} \gg \frac{\mathcal{L}(\mathcal{Y}_{L},\hat{\mathcal{Y}}_{L};\Theta_{L})}{\partial \Theta_L}.
\end{equation}
Proved.

\section{More Explanation of Energy Amplification Technique}\label{explanation_appdenix}
Our proposed energy amplification technique, which amplifies energy through spectrum flipping, has the following advantages:

\textbf{Simplicity}: It is straightforward to implement and requires no additional feature engineering.

\textbf{Equal Energy Levels}: Flipping the spectrum naturally allows low-energy components to attain the same energy levels as high-energy components, enabling the model to treat both equally.

\textbf{Adaptability}: The frequency-domain linear operation in Equation \ref{eq:freq_upsampling} not only adjusts the input length to match the prediction length but also fundamentally performs frequency-domain interpolation. This interpolation enables the model to learn amplitude scaling, dynamically adjusting the flipped spectrum adaptively for different datasets and forecasting scenarios.

\section{Supplementary Details of the Experiments}\label{results_appdenix}

\paragraph{\textbf{Experiment Settings.}} We follow the same data processing and train-validation-test set split protocol employed in iTransformer~\cite{liuitransformer}.
All the experiments are implemented in PyTorch 2.0.1 and conducted on a single NVIDIA RTX 3070 GPU with 8GB. 
We utilize ADAM~\cite{kingma2014adam} with an initial learning rate in $\left\{5\times 10^{-3},10^{-2},2\times 10^{-2} \right\}$ and L2 loss for the model optimization. The training epochs is fixed to 10 with batch size in $\left\{128, 256 \right\}$. The hidden size of FFNs is set from $\left\{128, 256, 512 \right\}$ with intermediate LeakyReLU activation function. The codes have been uploaded as supplementary and will be publicly available soon.


\paragraph{\textbf{Datasets.}} We evaluate the performance of our proposed Amplifier on eight popular datasets, including ETT, Electricity, Exchange, Traffic, and Weather datasets. 

The ETT~\cite{zhou2021informer} datasets contain two visions of the sub-dataset: ETTh and ETTm, which were collected from electricity transformers at intervals of 15 minutes and 1 hour, respectively, between July 2016 and July 2018. 

The Electricity~\cite{wu2021autoformer} dataset records the hourly electricity consumption of 321 clients from 2012 to 2014. 

The Exchange~\cite{lai2018modeling} dataset collects panel data of daily exchange rates of eight different countries including Singapore, Australia, British, Canada, Switzerland, China, Japan, and New Zealand ranging from 1990 to 2016.

The Traffic~\cite{sen2019think} dataset contains hourly traffic data measured by 862 sensors on San Francisco Bay area freeways since January 1, 2015.

The Weather~\cite{wu2021autoformer} dataset includes 21 meteorological factors collected every 10 minutes from the Weather Station of the Max Planck Biogeochemistry Institute in 2020. The data sampling interval is every 10 minutes.

The details of these datasets are presented in Table \ref{tab:dataset}.
\begin{table}[h]
    \centering
    {\fontsize{10pt}{10pt}\selectfont
    \scalebox{0.65}{
    \begin{tabular}{c c c c c c c c}
    \toprule[2pt]
       \multicolumn{2}{c}{Datasets}  &ETTm1(2) &ETTh1(2) &Electricity &Exchange  &Traffic &Weather \\
       \midrule
       \multicolumn{2}{c}{Channels} &7     &7       &321          &8        &862      &21 \\
       \multicolumn{2}{c}{Frequency} &15min  &Hourly  &Hourly  &Daily   &Hourly   &10min   \\
       \multicolumn{2}{c}{Timesteps} &69680   &17420   &26304   &7588    &17544   &52696      \\
       \multicolumn{2}{c}{Information} &Electricity &Electricity  &Electricity &Economy &Traffic &Weather  \\
    \bottomrule[2pt]
    \end{tabular}
    }}
    \caption{Summary of datasets.}
    \label{tab:dataset}
    \vspace{-3mm}
\end{table}

\paragraph{\textbf{Baselines.}} We choose ten well-acknowledged and state-of-the-art models for comparison to evaluate the effectiveness of our proposed Amplifier for time series forecasting, including MLP-based models, Frequency-based models, Transformer-based models, and TCN-based models. We introduce these models as bellow:

SparseTSF~\cite{lin2024sparsetsf} marks a significant milestone in advancing lightweight models for long-term time series forecasting, based on the Cross-Period Sparse Forecasting technique. Code is available at this repository: \url{https://github.com/lss-1138/SparseTSF}.

RLinear~\cite{li2023revisiting} employs linear mapping in long-term time series forecasting with RevIN (reversible normalization) and CI (Channel Independent) improve overall forecasting performance. Code is available at this repository: \url{https://github.com/plumprc/RTSF}.

DLinear~\cite{zeng2023transformers} utilizes a simple yet effective one-layer linear model to capture temporal relationships. Code is available at this repository: \url{https://github.com/cure-lab/LTSF-Linear}.

FreTS~\cite{yi2024frequency} explores a novel direction and make a new attempt to apply frequency-domain MLPs for time series forecasting, benefiting from global view and energy compaction. Code is available at this repository: \url{https://github.com/aikunyi/FreTS}.

FITS~\cite{xu2023fits} is a lightweight yet powerful model for time series analysis, essentially functioning as a low-pass filter. Code is available at this repository: \url{https://github.com/VEWOXIC/FITS}.

Fredformer~\cite{piao2024fredformer} addresses frequency bias in the Transformer architecture by introducing a framework that learns features equally across different frequency bands. Code is available at this repository: \url{https://github.com/chenzRG/Fredformer}.

iTransformer~\cite{liuitransformer} applies the attention and feed-forward network on the inverted dimensions and regards independent series as variate tokens. Code is available at this repository: \url{https://github.com/thuml/iTransformer}.

PatchTST~\cite{nietime} introduces an efficient design for Transformer-based models in time series forecasting by incorporating two essential components: patching and a channel-independent structure. Code is available at this repository: \url{https://github.com/yuqinie98/PatchTST}.

Stationary~\cite{liu2022non} proposes an effective approach to enhance series stationarity while updating the internal mechanism to reintegrate non-stationary information, thereby improving both data predictability and the model's predictive performance. Code is available at this repository: \url{https://github.com/thuml/Nonstationary_Transformers}.

TimesNet~\cite{wu2023timesnet} unravels complex temporal variations through a modular architecture, capturing both intraperiod and interperiod variations by converting the 1D time series into a collection of 2D tensors across multiple periods. Code is available at this repository: \url{https://github.com/thuml/TimesNet}.

\paragraph{\textbf{Full Results.}}The full multivariate time series forecasting results of Amplifier are presented in Table \ref{tab:app_input_96}, along with extensive evaluations of competitive counterparts. We compare extensive competitive models under different prediction lengths following the setting of iTransformer~\cite{liuitransformer} and Fredformer~\cite{piao2024fredformer}. For Amplifier, FreTS, and FITS we report the forecasting performance under five runs.
\begin{table*}[!b]
    \centering
    {\fontsize{10pt}{10pt}\selectfont
    \scalebox{0.68}{
    \begin{tabular}{l | c|c c|c c|c c| c c|c c |c c |c c |c c |c c |c c}
    \toprule[2pt]
       \multicolumn{2}{c|}{Models}  &\multicolumn{2}{c|}{\textbf{Amplifier}} &\multicolumn{2}{c|}{RLinear} &\multicolumn{2}{c|}{DLinear} &\multicolumn{2}{c|}{FreTS}  &\multicolumn{2}{c|}{FITS} &\multicolumn{2}{c|}{Fredformer} &\multicolumn{2}{c|}{iTransformer} &\multicolumn{2}{c|}{PatchTST}  &\multicolumn{2}{c|}{Stationary} &\multicolumn{2}{c}{TimesNet}  \\
       \cmidrule(r){1-2}\cmidrule(lr){3-4}\cmidrule(lr){5-6}\cmidrule(lr){7-8}\cmidrule(lr){9-10}\cmidrule(lr){11-12}\cmidrule(lr){13-14}\cmidrule(lr){15-16}\cmidrule(lr){17-18}\cmidrule(lr){19-20}\cmidrule(lr){21-22}
       \multicolumn{2}{c|}{Metrics}&MSE &MAE &MSE &MAE &MSE &MAE &MSE &MAE &MSE &MAE &MSE &MAE &MSE &MAE &MSE &MAE &MSE &MAE &MSE &MAE \\
       \midrule[1pt]
         \multirow{5}*{\rotatebox{90}{ETTm1}}
         & 96  &\textcolor{red}{\textbf{0.316}}  &\textcolor{red}{\textbf{0.355}}   &0.355  &0.376  &0.345  &0.372  &0.335  &0.372 &0.355  &0.375  &\textcolor{blue}{\textbf{0.326}} &\textcolor{blue}{\textbf{0.361}}  &0.334  &0.368  &0.329 &0.367  &0.386  &0.398  &0.338  &0.375   \\
         & 192 &\textcolor{red}{\textbf{0.361}}   &\textcolor{blue}{\textbf{0.381}}    &0.391  &0.392  &0.380  &0.389  &0.388  &0.401   &0.392  &0.393 &\textcolor{blue}{\textbf{0.363}} &\textcolor{red}{\textbf{0.380}}  &0.377  &0.391  &0.367 &0.385 &0.459  &0.444  &0.374  &0.387  \\
         & 336 &\textcolor{red}{\textbf{0.393}}   &\textcolor{red}{\textbf{0.402}}    &0.424  &0.415  &0.413  &0.413  &0.421  &0.426   &0.424  &0.414 &\textcolor{blue}{\textbf{0.395}} &\textcolor{blue}{\textbf{0.403}}  &0.426  &0.420  &0.399 &0.410  &0.495  &0.464   &0.410  &0.411 \\
         & 720 &\textcolor{blue}{\textbf{0.455}}  &\textcolor{blue}{\textbf{0.440}}   &0.487  &0.450  &0.474  &0.453  &0.486  &0.465   &0.487  & 0.449 &\textcolor{red}{\textbf{0.453}} &\textcolor{red}{\textbf{0.438}} &0.491  &0.459  &0.454  &0.439  &0.585  &0.516   &0.478  &0.450  \\
         \cmidrule(lr){2-22}
        &Avg &\textcolor{red}{\textbf{0.381}}  &\textcolor{red}{\textbf{0.394}}   &0.414 &0.407 &0.403 &0.407 &0.408  &0.416  &0.415  &0.408 &\textcolor{blue}{\textbf{0.384}} &\textcolor{blue}{\textbf{0.395}}  &0.407 &0.410 &0.387 &0.400 &0.481 &0.456 &0.400 &0.406 \\
         \midrule[1pt]
         \multirow{5}*{\rotatebox{90}{ETTm2}} 
         & 96  &\textcolor{red}{\textbf{0.176}}   &\textcolor{red}{\textbf{0.258}}   &0.182  &0.265  &0.193  &0.292  &0.189  &0.277 &0.183  &0.266 &\textcolor{blue}{\textbf{0.177}} &\textcolor{blue}{\textbf{0.259}}  &0.180  &0.264  &0.175  &0.259 &0.192  &0.274   &0.187  &0.267   \\
         & 192 &\textcolor{red}{\textbf{0.239}}  &\textcolor{red}{\textbf{0.300}}  &0.246  &0.304  &0.284  &0.362  &0.258  &0.326   &0.247  &0.305 & \textcolor{blue}{\textbf{0.241}} &\textcolor{red}{\textbf{0.300}} &0.250  &0.309  &0.241  &\textcolor{blue}{\textbf{0.302}}  &0.280  &0.339   &0.249  &0.309    \\
         & 336  &\textcolor{red}{\textbf{0.297}}  &\textcolor{red}{\textbf{0.338}} &0.307  &0.342  &0.369  &0.427  &0.343  &0.390  &0.307  &0.342 &\textcolor{blue}{\textbf{0.302}} &\textcolor{blue}{\textbf{0.340}} &0.311  &0.348  &0.305 &0.343  &0.334  &0.361    &0.321  &0.351   \\
         & 720  &\textcolor{red}{\textbf{0.393}}  &\textcolor{red}{\textbf{0.396}}   &0.407  &\textcolor{blue}{\textbf{0.398}}  &0.554  &0.522  &0.495  &0.480  &0.407  &0.399 &\textcolor{blue}{\textbf{0.397}} &\textcolor{red}{\textbf{0.396}} &0.412  &0.407  &0.402  &0.400  &0.417  & 0.413   &0.408  &0.403   \\
        \cmidrule(lr){2-22}
         &Avg &\textcolor{red}{\textbf{0.276}}  &\textcolor{red}{\textbf{0.323}}  &0.286 &0.327 &0.350 &0.401 &0.321  &0.368  &0.286  &0.328 &\textcolor{blue}{\textbf{0.279}} &\textcolor{blue}{\textbf{0.324}}   &0.288 &0.332 &0.281 &0.326 &0.306 &0.347 &0.291 &0.333 \\
         \midrule[1pt]
         \multirow{5}*{\rotatebox{90}{ETTh1}}
         & 96  &\textcolor{red}{\textbf{0.371}}  &\textcolor{red}{\textbf{0.392}}  &0.386  &\textcolor{blue}{\textbf{0.395}}  &0.386  &0.400  &0.395  &0.407  &0.386  &0.396 &\textcolor{blue}{\textbf{0.373}} &\textcolor{red}{\textbf{0.392}} &0.386  &0.405  &0.414  &0.419  &0.513  &0.491   &0.384  &0.402  \\
         & 192  &\textcolor{red}{\textbf{0.426}} &\textcolor{blue}{\textbf{0.422}}   &0.437  &0.424  &0.437  &0.432  &0.448  &0.440   &0.436  &0.423 &\textcolor{blue}{\textbf{0.433}} &\textcolor{red}{\textbf{0.420}}  &0.441  &0.436  &0.460  &0.445  &0.534  &0.504  &0.436  &0.429  \\
         & 336  &\textcolor{red}{\textbf{0.448}}   &\textcolor{red}{\textbf{0.434}}    &0.479  &0.446  &0.481  &0.459  &0.499  &0.472   &0.478 &0.444 &\textcolor{blue}{\textbf{0.470}} &\textcolor{blue}{\textbf{0.437}}  &0.487  &0.458  &0.501  &0.466  &0.588  &0.535  &0.491  &0.469  \\
         & 720  &\textcolor{blue}{\textbf{0.476}}  &\textcolor{blue}{\textbf{0.464}}   &0.481 &0.470  &0.519  &0.516  &0.558  &0.532  &0.502  &0.495 &\textcolor{red}{\textbf{0.467}} &\textcolor{red}{\textbf{0.456}}  &0.503  &0.491  &0.500  &0.488  &0.643  &0.616  &0.521  &0.500  \\
        \cmidrule(lr){2-22}
         &Avg &\textcolor{red}{\textbf{0.430}} &\textcolor{blue}{\textbf{0.428}}  &0.446 &0.434 &0.456 &0.452 &0.475  &0.463  &0.451  &0.440 &\textcolor{blue}{\textbf{0.435}} &\textcolor{red}{\textbf{0.426}}  &0.454 &0.447 &0.469 &0.454 &0.570 &0.537 &0.458 &0.450 \\
         \midrule[1pt]
         \multirow{5}*{\rotatebox{90}{ETTh2 }}
         & 96  &\textcolor{red}{\textbf{0.279}}  &\textcolor{red}{\textbf{0.337}}  &\textcolor{blue}{\textbf{0.288}}  &\textcolor{blue}{\textbf{0.338}}  &0.333  &0.387  &0.309 &0.364   &0.295  &0.350 &0.293 &0.342  &0.297  &0.349  &0.302  &0.348  &0.476  &0.458   &0.340  &0.374   \\
         & 192  &\textcolor{red}{\textbf{0.359}}  &\textcolor{red}{\textbf{0.389}} &0.374  &\textcolor{blue}{\textbf{0.390}}  &0.477  &0.476  &0.395 &0.425  &0.381  &0.396  &\textcolor{blue}{\textbf{0.371}} &\textcolor{red}{\textbf{0.389}}  &0.380  &0.400  &0.388  &0.400  &0.512  &0.493   &0.402  &0.414  \\
         & 336  &\textcolor{red}{\textbf{0.377}}   &\textcolor{red}{\textbf{0.406}}   &0.415  &0.426  &0.594  &0.541  &0.462 &0.467  &0.426  &0.438 &\textcolor{blue}{\textbf{0.382}} &\textcolor{blue}{\textbf{0.409}}  &0.428  &0.432  &0.426  &0.433  &0.552  &0.551   &0.452  &0.452  \\
         & 720  &\textcolor{blue}{\textbf{0.420}}   &\textcolor{red}{\textbf{0.432}}  &0.420  &0.440  &0.831  &0.657  &0.721 &0.604  &0.431  &0.446 &\textcolor{red}{\textbf{0.415}} &\textcolor{blue}{\textbf{0.434}}  &0.427  &0.445  &0.431  &0.446  &0.562  &0.560   &0.462  &0.468  \\
        \cmidrule(lr){2-22}
         &Avg &\textcolor{red}{\textbf{0.359}} &\textcolor{red}{\textbf{0.391}} &0.374 &0.398 &0.559 &0.515 &0.472  &0.465  &0.383  &0.408 &\textcolor{blue}{\textbf{0.365}} &\textcolor{blue}{\textbf{0.393}}  &0.383 &0.407 &0.387 &0.407 &0.526 &0.516 &0.414 &0.427 \\
         \midrule[1pt]
         \multirow{5}*{\rotatebox{90}{ECL}} 
         & 96  &\textcolor{red}{\textbf{0.147}}  &0.243  &0.201  &0.281  &0.197  &0.282  &0.176  &0.258  &0.200  &0.278 &\textcolor{red}{\textbf{0.147}} &\textcolor{blue}{\textbf{0.241}} &\textcolor{blue}{\textbf{0.148}}  &\textcolor{red}{\textbf{0.240}}  &0.181  &0.270  &0.169  &0.273   &0.168  &0.272   \\
         & 192  &\textcolor{red}{\textbf{0.157}}  &\textcolor{red}{\textbf{0.251}}  &0.201  &0.283  &0.196  &0.285  &0.175  &0.262  &0.200  & 0.280  &0.165 &0.258 &\textcolor{blue}{\textbf{0.162}}  &\textcolor{blue}{\textbf{0.253}}  &0.188  &0.274  &0.182  &0.286   &0.184  &0.289  \\
         & 336  &\textcolor{red}{\textbf{0.174}}  &\textcolor{red}{\textbf{0.269}}  &0.215  &0.298  &0.209  &0.301  &0.185  &0.278  &0.214  & 0.295 &\textcolor{blue}{\textbf{0.177}} &\textcolor{blue}{\textbf{0.273}}  &0.178  &\textcolor{red}{\textbf{0.269}}  &0.204  &0.293  &0.200  &0.304   &0.198  &0.300  \\
         & 720  &\textcolor{red}{\textbf{0.206}}  &\textcolor{red}{\textbf{0.296}}  &0.257  &0.331  &0.245  &0.333  &0.220  &0.315  &0.255  & 0.327 &\textcolor{blue}{\textbf{0.213}} &\textcolor{blue}{\textbf{0.304}}  &0.225  &0.317  &0.246  &0.324  &0.222  &0.321   &0.220  &0.320  \\
        \cmidrule(lr){2-22}
         &Avg &\textcolor{red}{\textbf{0.171}}  &\textcolor{red}{\textbf{0.265}}  &0.219 &0.298 &0.212 &0.300 &0.189  &0.278  &0.217  &0.295 &\textcolor{blue}{\textbf{0.175}}  &\textcolor{blue}{\textbf{0.269}}  &0.178 &0.270 &0.216 &0.304 &0.193 &0.296 &0.192 &0.295 \\
         \midrule[1pt]
         \multirow{5}*{\rotatebox{90}{Exchange}} 
         & 96  &\textcolor{red}{\textbf{0.083}}  &\textcolor{red}{\textbf{0.202}}  &0.093  &0.217  &0.088  &0.218  &0.091  &0.217  &\textcolor{blue}{\textbf{0.084}} &\textcolor{blue}{\textbf{0.203}} &\textcolor{blue}{\textbf{0.084}}  &\textcolor{red}{\textbf{0.202}}   &0.086  &0.206  &0.088  &0.205  &0.111  & 0.237  &0.107  &0.234  \\
         & 192  &\textcolor{red}{\textbf{0.175}}  &\textcolor{red}{\textbf{0.297}}  &0.184  &0.307  &\textcolor{blue}{\textbf{0.176}}  &0.315  &0.175  &0.310 &0.177  &\textcolor{blue}{\textbf{0.298}}  &0.183  &0.302   &0.177  &0.299  &\textcolor{blue}{\textbf{0.176}}  &0.299  &0.219  &0.335   &0.226  &0.344  \\
         & 336  &0.328   &0.414   &0.351  &0.432  &\textcolor{blue}{\textbf{0.313}}  &0.427  &0.334  &0.434 &0.321  &\textcolor{blue}{\textbf{0.410}} &0.335  &0.418   &0.331  &0.417  &\textcolor{red}{\textbf{0.301}}  &\textcolor{red}{\textbf{0.397}}  &0.421  &0.476  &0.367  &0.448   \\
         & 720  &0.858  &0.696   &0.886  &0.714  &\textcolor{blue}{\textbf{0.839}}  &0.695  &\textcolor{red}{\textbf{0.716}}  &\textcolor{red}{\textbf{0.674}}  &0.828  &\textcolor{blue}{\textbf{0.685}}  &0.893  &0.711   &0.847  &0.691  &0.901  &0.714 &1.092  &0.769   &0.964  &0.746  \\
        \cmidrule(lr){2-22}
         &Avg  &0.361 &\textcolor{blue}{\textbf{0.402}} &0.378 &0.417 &0.354 &0.414 &\textcolor{red}{\textbf{0.329}}  &0.409  &\textcolor{blue}{\textbf{0.353}}  &\textcolor{red}{\textbf{0.399}} &0.374  &0.408   &0.360 &0.403 &0.367 &0.404 &0.461 &0.454 &0.416 &0.443 \\
         \midrule[1pt]
         \multirow{5}*{\rotatebox{90}{Traffic}} 
         & 96  &0.455  &0.298  &0.649  &0.389  &0.650  &0.396  &0.593 &0.378  &0.651  & 0.391 &\textcolor{blue}{\textbf{0.406}} &\textcolor{blue}{\textbf{0.277}} &\textcolor{red}{\textbf{0.395}}  &\textcolor{red}{\textbf{0.268}}  &0.462  &0.295   &0.612  &0.338  &0.593  &0.321 \\
         & 192 &0.470  &0.316  &0.601  &0.366  &0.598  &0.370  &0.595 &0.377  &0.602  &0.363 &\textcolor{blue}{\textbf{0.426}} &\textcolor{blue}{\textbf{0.290}}  &\textcolor{red}{\textbf{0.417}}  &\textcolor{red}{\textbf{0.276}}  &0.466  &0.296   &0.613  &0.340  &0.617  &0.336  \\
         & 336 &0.479  &0.316  &0.609  &0.369  &0.605  &0.373  &0.609 &0.385  &0.609  &0.366 &\textcolor{red}{\textbf{0.432}} &\textcolor{red}{\textbf{0.281}}  &\textcolor{blue}{\textbf{0.433}}  &\textcolor{blue}{\textbf{0.283}}  &0.482  &0.304   &0.618  &0.328  &0.629  &0.336 \\
         & 720 &0.523  &0.328  &0.647  &0.387  &0.645  &0.394  &0.673 &0.418   &0.647  &0.385 &\textcolor{red}{\textbf{0.463}} &\textcolor{red}{\textbf{0.300}} &\textcolor{blue}{\textbf{0.467}}  &\textcolor{blue}{\textbf{0.302}}  &0.514  &0.322   &0.653  &0.355  &0.640  &0.350 \\
        \cmidrule(lr){2-22}
         &Avg  &0.482  &0.315  &0.626 &0.378 &0.625 &0.383 &0.618  &0.390  &0.627  &0.376 &\textcolor{blue}{\textbf{0.431}} &\textcolor{blue}{\textbf{0.287}}  &\textcolor{red}{\textbf{0.428}} &\textcolor{red}{\textbf{0.282}} &0.481 &0.304 &0.624 &0.340 &0.620 &0.336 \\
         \midrule[1pt]
         \multirow{5}*{\rotatebox{90}{Weather}} 
         & 96  &\textcolor{red}{\textbf{0.156}}  &\textcolor{red}{\textbf{0.204}} &0.192  &0.232  &0.196  &0.255  &0.174  &0.208  &0.166  &0.213 &\textcolor{blue}{\textbf{0.163}} &\textcolor{blue}{\textbf{0.207}}  &0.174  &0.214  &0.177  &0.218  &0.173  &0.223   &0.172  &0.220   \\
         & 192  &\textcolor{red}{\textbf{0.209}}  &\textcolor{red}{\textbf{0.249}}  &0.240  &0.271  &0.237  &0.296  &0.219  &\textcolor{blue}{\textbf{0.250}}  &0.213  &0.254 &\textcolor{blue}{\textbf{0.211}} &0.251 &0.221  &0.254  &0.225  &0.259   &0.245  &0.285   &0.219  &0.261 \\
         & 336  &\textcolor{red}{\textbf{0.264}}  &\textcolor{red}{\textbf{0.290}}  &0.292  &0.307  &0.283  &0.335  &0.273  &\textcolor{red}{\textbf{0.290}}  &0.269  &0.294 &\textcolor{blue}{\textbf{0.267}} &\textcolor{blue}{\textbf{0.292}} &0.278  &0.296  &0.278  &0.297  &0.321  &0.338   &0.280  &0.306   \\
         & 720  &\textcolor{blue}{\textbf{0.343}}  &0.342  &0.364  &0.353  &0.345  &0.381  &\textcolor{red}{\textbf{0.334}}  &\textcolor{red}{\textbf{0.332}}  &0.346  &0.343 &\textcolor{blue}{\textbf{0.343}} &\textcolor{blue}{\textbf{0.341}} &0.358  &0.347  &0.354  &0.348  &0.414  &0.410   &0.365  &0.359  \\
        \cmidrule(lr){2-22}
         &Avg  &\textcolor{red}{\textbf{0.243}}   &\textcolor{red}{\textbf{0.271}}  &0.272 &0.291 &0.265 &0.317 &0.250  &0.270  &0.249  &0.276 &\textcolor{blue}{\textbf{0.246}} &\textcolor{blue}{\textbf{0.272}}  &0.258 &0.278 &0.259 &0.281 &0.288 &0.314 &0.259 &0.287 \\
        \midrule[1pt]
        \multicolumn{2}{c|}{$1^{st}$ Count} &\textcolor{red}{\textbf{28}} &\textcolor{red}{\textbf{25}} &0 &0 &0 &0 &3 &3 &0 &1 &\textcolor{blue}{\textbf{6}} &\textcolor{blue}{\textbf{12}}  &3 &5 &1 &1 &0 &0 &0 &0 \\
         \bottomrule[2pt]
    \end{tabular}
    }}
    \caption{Full results of eight datasets, with the best results are in \textcolor{red}{red} and the second best are \textcolor{blue}{blue}. We set the lookback window size $L$ as 96 and the prediction length as $\tau \in \{96, 192, 336, 720\}$. Avg means the average results from all four prediction lengths.}
    \label{tab:app_input_96}
    \vspace{-2mm}
\end{table*}

\end{document}